\documentclass[10pt,twocolumn,letterpaper]{article}

\usepackage[pagenumbers]{cvpr} 

\usepackage{makecell}

\definecolor{cvprblue}{rgb}{0.21,0.49,0.74}
\usepackage[pagebackref,breaklinks,colorlinks,allcolors=cvprblue]{hyperref}

\def\paperID{17} 
\def\confName{CVPR}
\def\confYear{2026}

\title{Which Reconstruction Model Should a Robot Use? Routing Image-to-3D Models for Cost-Aware Robotic Manipulation}

\author{Akash Anand
\and
Aditya Agarwal\\
Massachusetts Institute of Technology\\
{\tt\small \{akash10, adityaag, lpk\}@mit.edu}
\and
Leslie Pack Kaelbling
}

\begin{document}
\maketitle
\begin{abstract}
Robotic manipulation tasks require 3D mesh reconstructions of varying quality: dexterous manipulation demands fine-grained surface detail, while collision-free planning tolerates coarser representations. Multiple reconstruction methods offer different cost-quality tradeoffs, from Image-to-3D models---whose output quality depends heavily on the input viewpoint---to view-invariant methods such as structured light scanning. Querying all models is computationally prohibitive, motivating per-input model selection. We propose SCOUT, a novel routing framework that decouples reconstruction scores into two components: (1) the relative performance of viewpoint-dependent models, captured by a learned probability distribution $\hat{\mathbf{p}}$, and (2) the overall image difficulty, captured by a scalar partition function estimate $\hat{z}$. As the learned network operates only over the viewpoint-dependent models, view-invariant pipelines can be added, removed, or reconfigured without retraining. SCOUT also supports arbitrary cost constraints at inference time, accommodating the multi-dimensional cost constraints common in robotics. We evaluate on the Google Scanned Objects, BigBIRD, and YCB datasets under multiple mesh quality metrics, demonstrating consistent improvements over routing baselines adapted from the LLM literature across various cost constraints. We further validate the framework through robotic grasping and dexterous manipulation experiments. We release the code and additional results on our \href{https://scout-model-routing.github.io}{website}.
\end{abstract}    
\section{Introduction}
\label{sec:intro}

\begin{figure*}[t]
    \centering
    \includegraphics[width=\textwidth]{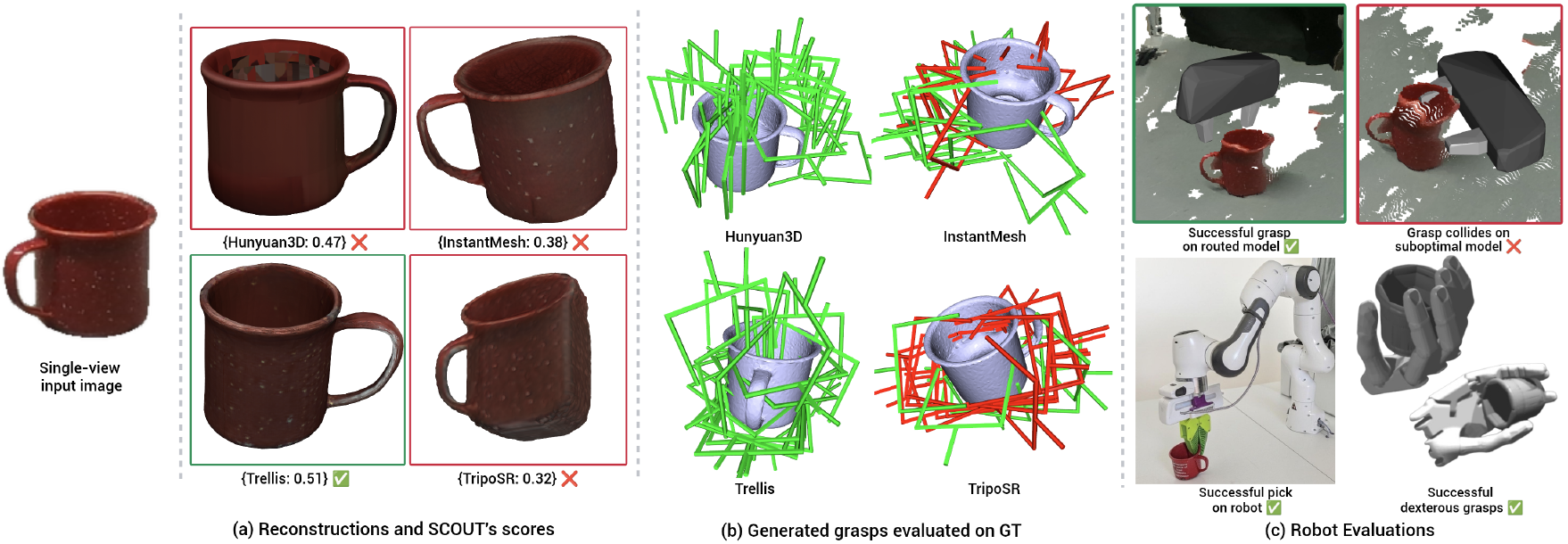}
    \caption{Overview of SCOUT. Given an input image, (a) SCOUT routes to the most suitable reconstruction model and generates the selected 3D reconstruction (compared against other candidate models). (b) Grasp proposals generated on the reconstructed mesh, evaluated on the ground-truth mesh; colliding grasps are shown in red. (c) Utility of SCOUT's reconstruction-aware routing in downstream robust robot grasping and dexterous manipulation.}
    \label{fig:myimage}
    \vspace{-14px}
\end{figure*}

Single-image 3D reconstruction has advanced rapidly in recent years. Early optimization-based approaches demonstrated that strong generative priors~\cite{liu2023zero1to3, shi2023zero123++} could lift a single image into 3D, but required costly per-object optimization~\cite{poole2022dreamfusion, wang2023score,3dfuse, qian2023magic123} taking minutes to hours, limiting their use in real-time robotics applications. Recent Image-to-3D models remove this bottleneck, producing mesh reconstructions in seconds without per-object optimization. These models span diverse architectures---feed-forward prediction~\cite{tochilkin2024triposrfast3dobject,hong2023lrm}, multi-view diffusion followed by reconstruction~\cite{xu2024instantmeshefficient3dmesh,liu2023one2345}, and iterative generation in learned 3D latent spaces~\cite{xiang2024structuredtrellis, hunyuan3d22025tencent}---and offer complementary strengths: some models prioritize inference speed while others emphasize geometric fidelity or texture quality, and their relative performance varies across object categories and input viewpoints. As a result, no single method dominates.

This is relevant for robotic manipulation, where a robot often has access to only a single initial view of an object~\cite{nolte2025single,sen2023scarp}, and acquiring additional
viewpoints is time-consuming and not always feasible~\cite{lundell2020beyond, mohammadi20233dsgrasp}. Several works use Image-to-3D models for perception in downstream manipulation~\cite{scenecomplete, nolte2025single,kashyap2025single,iwase2025zerograsp,iwase2024octmae}. However, these works fix the choice of Image-to-3D model, forgoing potential gains from selecting the best model per input. Moreover, whether a given reconstruction is sufficient depends on the task: dexterous manipulation demands fine-grained surface detail, whereas collision-free motion planning tolerates coarser representations. Beyond Image-to-3D models, methods such as structured light scanning provide reconstructions whose quality depends on the object and hardware configuration rather than on the robot's initial viewpoint, but they require additional sensing time. We refer to such methods as \textit{view-invariant}, in contrast to the \textit{viewpoint-dependent} Image-to-3D models above. The choice of reconstruction method therefore depends on the input viewpoint, task requirements, and available budget.

We address this problem via model routing: selecting a reconstruction model for a given input prior to inference. The routing decision balances reconstruction accuracy against user-specified costs---including latency, memory, and monetary expenses. While model routing has emerged as an active research area for LLMs~\cite{lu2024routing, chen2024routerdc, hu2024routerbench, ong2024routellm, li2025rethinking}, it has not been explored for 3D reconstruction. The two settings differ in important ways: the costs of reconstruction models are largely input-independent, robotics applications require multi-dimensional cost coefficient vectors rather than a single scalar tradeoff, and the model pool includes both viewpoint-dependent methods (\eg, Image-to-3D models) and view-invariant methods that can be reconfigured post-deployment, such as by changing the number of views captured.

We propose \textbf{SCOUT} (\textbf{\underline{S}core-\underline{C}onditioned \underline{O}ptimal \underline{U}tility \underline{T}argeting}), a routing framework that decouples reconstruction scores into two components: (1) the relative performance of viewpoint-dependent models, captured by a learned probability distribution, and (2) the overall difficulty of the input image, captured by a scalar partition function estimate. This decoupling stabilizes training and provides a key scalability advantage: the learned network predicts only over the viewpoint-dependent models, so its size and training procedure are entirely independent of the view-invariant methods. SCOUT also supports arbitrary cost coefficient vectors at inference time without retraining.

We evaluate SCOUT on the Google Scanned Objects~\cite{downs2022google}, BigBIRD~\cite{singh2014bigbird}, and YCB~\cite{calli2015ycb} datasets under multiple mesh quality metrics, including Density-aware Chamfer Distance (DCD)~\cite{wu2021density}, IoU, and geometric metrics from Eval3D~\cite{duggal2025eval3d}. SCOUT consistently outperforms routing baselines adapted from the LLM literature across diverse cost coefficient vector subspaces, and we validate its practical utility through robotic manipulation experiments, including collision-free grasp proposal evaluation, dexterous manipulation in simulation, and real-world pick-and-place on a Franka Panda robot.

The main contributions of this work are: (1) the first formulation of model routing for 3D reconstruction, accommodating viewpoint-dependent and view-invariant methods under arbitrary cost constraints; (2) a theoretically justified partition function proxy that recovers absolute scores from the learned relative distribution with provably optimal weighting; and (3) a decoupling procedure separating image difficulty from relative model performance, enabling view-invariant methods to be added without retraining.

\section{Related Work}
\label{sec:related_work}

\textbf{Mesh reconstruction models.} Early single-image 3D reconstruction methods relied on per-object optimization, using score distillation sampling~\cite{poole2022dreamfusion,wang2023score} or novel view synthesis models~\cite{liu2023zero1to3, shi2023zero123++} to iteratively optimize a 3D representation from a single image~\cite{qian2023magic123,3dfuse}. Subsequent methods replaced per-object optimization with fast inference~\cite{hong2023lrm, liu2023one2345, liu2024one2345++}. Among recent and widely used Image-to-3D models, TripoSR~\cite{tochilkin2024triposrfast3dobject} prioritizes low-latency inference via a feed-forward transformer, Hunyuan3D~\cite{hunyuan3d22025tencent} emphasizes high-quality texture generation through a two-stage multi-view diffusion and reconstruction pipeline, InstantMesh~\cite{xu2024instantmeshefficient3dmesh} focuses on geometrically plausible reconstructions with accurate depths and surface normals, and TRELLIS~\cite{xiang2024structuredtrellis} employs a unified structured latent representation that captures intricate geometric detail. However, these models exhibit complementary failure modes, with performance varying across object categories and viewpoints. The diversity of available reconstruction pipelines motivates the need for a principled method to select the most appropriate reconstruction model given a specific input and user-defined cost constraints.

\textbf{Routing models.} Querying all models for each input is computationally intractable, motivating a routing network to select a single reconstruction model per input. To the best of our knowledge, routing has not been explored for 3D reconstruction, though it has emerged as an active research area for LLMs. ZOOTER~\cite{lu2024routing} trains a router via distillation from a reward model, and RouterDC~\cite{chen2024routerdc} learns per-model embeddings using contrastive losses. While both achieve strong routing performance, neither supports adjusting the cost-quality tradeoff at inference time without retraining.

Recognizing this limitation, RouterBench~\cite{hu2024routerbench} introduces baselines and metrics for evaluating routing models that can operate across varying cost-quality tradeoffs without retraining. Li~\cite{li2025rethinking} analyzes the proposed baselines and additional architectures, finding that simple models such as kNNs and linear regressors consistently outperform more complex neural approaches.
\section{Methods}
\label{sec:methods}

In this section, we first formalize the model routing problem for the 3D reconstruction domain and then introduce SCOUT, a routing framework that decouples reconstruction scores into the relative performance of viewpoint-dependent models conditioned on the input image and a scalar shift term capturing the overall difficulty of the input image.

\subsection{Problem formulation}
\label{sec:problem_formulation}
Let $\mathcal{M} = \{m_1, \cdots, m_k\}$ denote a set of $k$ candidate 3D reconstruction models, each producing a mesh from an input image of the current robot viewpoint, $x$. We seek a router that selects the optimal model $m^*$ for input $x$, determined by: a score function $s(x, m)$ that measures reconstruction quality (higher is better), and an image-independent cost function $c(m)$ that captures a user-specified penalization of each model. Given a training dataset $\mathcal{D}_{train} = \{\left(x^{(1)}, \mathbf{s}^{(1)}\right), \cdots, \left(x^{(n)}, \mathbf{s}^{(n)}\right)\}$, where $\mathbf{s}^{(i)} \in \mathbb{R}^k$ denotes the vector of reconstruction scores $s(x^{(i)}, m_j)$ for $j \in [k]$, we aim to learn a router $h$ that takes an input image $x$ and an arbitrary cost coefficient vector $[c(m_1),\cdots, c(m_k)] \equiv\mathbf{c} \in \mathbb{R}^k$, and outputs a model selection $\hat{m}$ to maximize $s(x, \hat{m}) - c(\hat{m})$. 

In the LLM setting, prior work~\cite{li2025rethinking, jitkrittum2025universal} defines the most general routing problem as maximizing the following objective for a user-defined scalar $\lambda$:
\begin{equation}
    m^* = \arg\max_{m \in \mathcal{M}} s(x, m) - \lambda \cdot c(x, m).
\end{equation}
Our formulation differs in three aspects: the dependence of $c$ on $m$ rather than on both $x$ and $m$, the use of an arbitrary cost coefficient vector $\mathbf{c} \in \mathbb{R}^k$ rather than a scalar-weighted cost $\lambda \cdot \mathbf{c_{\mathrm{fixed}}}$, and the composition of the model pool $\mathcal{M}$, which in our setting includes both viewpoint-dependent Image-to-3D models and view-invariant methods. We expand on these differences in turn.

First, in the LLM setting, inference cost is a function of the number of input and output tokens~\cite{pan2025cost}, both of which depend on the input prompt. Therefore, the cost function depends on both the input prompt and the queried model. In contrast, 3D reconstruction models often employ architectures with fixed computational graphs (\eg, pure feed-forward~\cite{tochilkin2024triposrfast3dobject} or fixed-step diffusion pipelines~\cite{xu2024instantmeshefficient3dmesh}), resulting in a constant inference cost. Even when this does not hold exactly, the approximation $c(x, m)\approx c(m)$ remains valid in practice and is adopted in several LLM routing works~\cite{hu2024routerbench, li2025llm}.

Second, as the LLM routing problem originated from selecting between paid API endpoints, cost is expressed in dollars per API call. In contrast, routing between 3D reconstruction models for robotics applications requires a more task-specific cost structure. The cost may simultaneously reflect both memory footprint and inference latency, spanning two dimensions rather than the one-dimensional family $\{\lambda \cdot \mathbf{c}_{\mathrm{fixed}} :\lambda \in \mathbb{R}\}$ traced by sweeping a single scalar $\lambda$ over a fixed cost vector. The cost structure can be considerably more complex in practice: a robot may assign a cost of $\infty$ to models exceeding its memory budget, yielding a binary feasibility constraint. We therefore argue that a routing framework must support an arbitrary cost coefficient vector $\mathbf{c} \in \mathbb{R}^k$, rather than a single scalar tradeoff parameter.

Third, unlike LLM routing, which considers only generative models, 3D reconstruction includes two fundamentally different pipeline classes. Viewpoint-dependent models, such as Image-to-3D models, produce scores that depend on both the input image and the model---a frontal (face-on) view can yield far worse reconstructions than a corner view. View-invariant methods, such as laser scanning and structured light scanning, capture their own viewpoints, so their reconstruction quality is largely independent of the robot's initial view. The quality instead depends on object-level properties such as texture and scanner configurations (\eg, the number of captured views and sensor resolution).

This distinction has two implications for routing. First, because view-invariant methods are not conditioned on the input view, including them in a joint learned model alongside viewpoint-dependent methods is both unnecessary and harmful. It is unnecessary because their scores can be estimated without a learned image-conditioned network---for example, via texture analysis, object priors, or as a function of the scanner configurations. It is harmful because the training objective for viewpoint-dependent models benefits from reasoning about \textit{relative} performance across models for a given input (\cref{sec:scout}), and mixing in scores that do not vary with the input view corrupts this relative signal. Second, the number of view-invariant configurations can be large---different scanners, view counts, and hardware setups each constitute a distinct option with its own cost-quality tradeoff---and a robot's available hardware may change between deployments. Coupling these into the learned network would require retraining whenever a new configuration is added and degrades performance as the number of configurations grows, as shown in \cref{fig:number_of_fixed}. Handling them separately avoids both problems.

In our experiments, we approximate view-invariant methods with fixed scores $s(m)$ that depend only on the method. This approximation is reasonable for our datasets with diffuse textures~\cite{downs2022google}, but in general $s(m)$ can be replaced with many estimation techniques, such as those listed above.

\subsection{SCOUT}
\label{sec:scout}
SCOUT is designed for two properties: (1) \textit{scalability} — view-invariant methods can be added or reconfigured without retraining; and (2) \textit{flexibility} — routing under arbitrary $\mathbf{c} \in \mathbb{R}^k$. The framework consists of three stages: preprocessing, score decoupling, and utility optimization.

\textbf{Preprocessing.} Let $\mathbf{s}^{(i)} = (s_1^{(i)}, \cdots, s_k^{(i)})$ denote the vector of reconstruction scores for all $k$ models on input $x^{(i)}$. We partition this vector into scores for the $k_1$ viewpoint-dependent models, and scores for the $k_2$ view-invariant models, such that $\mathbf{s}^{(i)} \equiv \left[\mathbf{s}_{\mathrm{dep}}^{(i)}, \mathbf{s}_{\mathrm{inv}}^{(i)}\right]$ with $\mathbf{s}_{\mathrm{dep}}^{(i)} \in \mathbb{R}^{k_1}$, $\mathbf{s}_{\mathrm{inv}}^{(i)} \in \mathbb{R}^{k_2}$, and $k_1 + k_2 = k$.

Following ZOOTER~\cite{lu2024routing} and motivated by the variance in reconstruction quality across object categories, we apply tag-based score smoothing to $\mathbf{s}_{\mathrm{dep}}^{(i)}$ by grouping objects into semantic categories and computing category-averaged scores. We denote this grouping as a tagger $\mathcal{T}: \mathcal{X} \rightarrow \mathrm{tag}$, and define the tag-averaged scores $\mathbf{s}_{\mathcal{T}(x^{(i)})}$ as the mean of $\mathbf{s}_{\mathrm{dep}}^{(i)}$ over all training examples assigned to the same tag as $x^{(i)}$. The smoothed scores are then given by:
\begin{equation}
    \tilde{\mathbf{s}}_{\mathrm{dep}}^{(i)} = \beta \mathbf{s}^{(i)}_{\mathrm{dep}} + (1-\beta)\mathbf{s}_{\mathcal{T}(x^{(i)})},
\end{equation}
where $\beta \in [0,1]$ is a hyperparameter controlling the degree of smoothing. Smoothing is applied only when learning relative model performance, as described below; the final recovered scores target the original unsmoothed values $\mathbf{s}_{\mathrm{dep}}^{(i)}$, which are needed for comparison with $\mathbf{s}_{\mathrm{inv}}^{(i)}$.

\textbf{Learned models.} The smoothed scores $\tilde{\mathbf{s}}_{\mathrm{dep}}^{(i)}$ can be predicted using several possible training objectives, including MSE directly on the smoothed scores, or KL divergence on the softmax distribution of the smoothed scores. We adopt the KL divergence formulation for the following reason: an easy input image (\eg, a corner view) may yield uniformly high scores across all viewpoint-dependent models, while a hard image (\eg, frontal) may yield uniformly low scores. A direct regression objective such as MSE treats these absolute score values as the learning target, whereas KL divergence operates on the relative performance of the viewpoint-dependent models by normalizing out image difficulty. This is precisely why decoupling is necessary: KL divergence factors out image difficulty only when all models in the softmax share the same difficulty-induced offset, which holds for viewpoint-dependent models but not for view-invariant methods. For each input image, we convert the smoothed scores $\tilde{\mathbf{s}}_{\mathrm{dep}}^{(i)}$ into a probability distribution via the softmax with temperature $T$:
\begin{equation}
    {p}^{(i)}_j = \frac{e^{\tilde{{s}}_{\mathrm{dep}, j}^{(i)}/T}}{\sum_{m=1}^{k_1}e^{\tilde{{s}}_{\mathrm{dep}, m}^{(i)}/T}},
\end{equation}
and train a neural network to map images $x$ into predictions $\hat{\mathbf{p}} \in \mathbb{R}^{k_1}$ by minimizing the KL divergence.

Our approach decouples the predicted scores into two interpretable components: the relative performance of the viewpoint-dependent models, captured by $\hat{\mathbf{p}}$, and the overall difficulty of the input image, captured by a scalar estimate $\hat{z}$. Importantly, $\hat{z}$ applies a uniform offset to all viewpoint-dependent scores, preserving their relative utilities while enabling direct comparison with view-invariant scores $\mathbf{s}_{\mathrm{inv}}$.

Recovering the unsmoothed scores from $\hat{\mathbf{p}}$ requires estimating a partition function. We define
$z^{(i)}=\sum_{m=1}^{k_1}e^{{{s}}_{\mathrm{dep}, m}^{(i)}/T}$
over the original unsmoothed scores, so that $T \cdot \ln(\hat{p}_j^{(i)}\cdot z^{(i)})$ approximates $s^{(i)}_{\mathrm{dep},j}$. If only viewpoint-dependent models were present, $T \cdot \ln(\hat{\mathbf{p}})$ could be used directly for utility optimization, as all scores would be offset by the same constant $T\cdot\ln(z^{(i)})$. However, since the viewpoint-dependent scores must be compared against the view-invariant scores $\mathbf{s}_{\mathrm{inv}}$, an estimate $\hat{z}$ is required to recover absolute score values.

We learn $\hat{z}$ via a second supervised model that takes $x$ as input.  Because the scores are noisy, we use per-model proxies
\begin{equation}
    \tilde{z}_j^{(i)} = \frac{e^{{{s}}_{\mathrm{dep}, j}^{(i)}/T}}{{\hat{p}}^{(i)}_j}
\end{equation}
rather than using $z^{(i)}$ directly. Note that unsmoothed scores appear in the proxy: smoothing blends per-image scores toward category averages, suppressing the per-image variation that $z$ must capture (see \cref{sec:smoothing_approximation}).  This yields $k_1$ proxies for $z^{(i)}$, which must be combined into a single estimate since all viewpoint-dependent scores are offset by the same constant $T \cdot \ln(z)$. As derived in \cref{sec:derivation}, the optimal combination can be approximated by a weighted average:
\begin{equation}
    \tilde{z}^{(i)} = \sum_{j=1}^{k_1} w_j\tilde{z}_j^{(i)}, \space  w_j \propto \frac{1}{S_{j}^2(1-R_j^2)\mathbb{E}_i\!\!\left[\dfrac{z^4}{e^{2s_{\mathrm{dep},j}/T}}\right]}
\end{equation}
where $R_j^2$ is the coefficient of determination of the predicted probabilities $\hat{{p}}_j$ relative to the true values ${p}_j$ for model $m_j$ and $S_{j}^2=\operatorname{Var}[\hat{{p}}_j]$. The resulting $\tilde{z}^{(i)}$ values serve as regression targets for the second supervised model. We compare this weighting against three alternatives: the ground-truth $z^{(i)}$, equal weighting, and one-hot weighting ($\tilde{z}^{(i)}=\tilde{z}_{j^*}^{(i)} \text{ where } j^*=\arg\max_jw_j$), and show improved performance using a one-sided t-test (see \cref{sec:weighted_proxy_results}).

\textbf{Utility optimization.} Given the predicted probability distribution $\hat{\mathbf{p}}$ and the estimated partition function $\hat{z}$, the optimal reconstruction method is selected as $\hat{m} = \arg \max_{m_j \in \mathcal{M}} \hat{s}(x, m_j) - \mathbf{c}[j]$, where the estimated score is defined as:
\begin{equation}
\hat{s}(x, m_j) =
\begin{cases}
T \cdot \ln\left(\hat{{p}}_j \cdot \hat{z}\right),
& \text{if } j \le k_1, \\[6pt]
s_{\mathrm{inv}}(m_j),
& \text{if } k_1 < j \le k.
\end{cases}
\end{equation}

\begin{figure}
    \centering
    \includegraphics[width=0.8\linewidth]{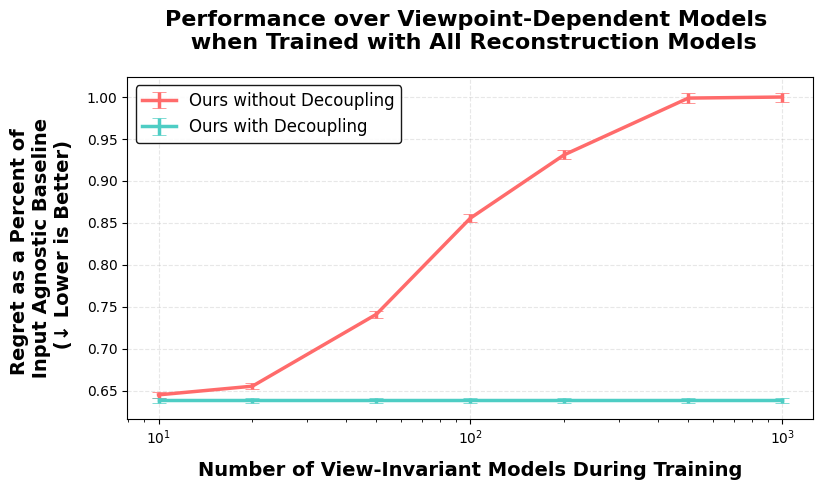}
    \caption{Effect of the number of view-invariant methods on routing performance when evaluated on viewpoint-dependent methods. With decoupling, regret over viewpoint-dependent models remains constant regardless of how many view-invariant methods are included during training. Without decoupling, regret increases as the number of view-invariant methods grows.}
    \label{fig:number_of_fixed}
    \vspace{-16px}
\end{figure}

For view-invariant methods, $s_{\mathrm{inv}}(m_j)$ may depend on object-level properties, such as texture or reflectance, but it does not depend on the viewpoint from which the robot initially observes the object. Because the learned network only predicts $\hat{\mathbf{p}}$ over the $k_1$ viewpoint-dependent models, new view-invariant configurations can be added, removed, or reconfigured at inference time without retraining. In contrast, applying the softmax over the full score vector $\mathbf{s}^{(i)}$, including view-invariant scores, would couple the learned distribution to $k_2$, resulting in degraded performance as the number of view-invariant methods increases, as demonstrated in \cref{fig:number_of_fixed}.

\section{Experiments}
\label{sec:experiments}

In this section, we describe the experimental setup, evaluate SCOUT on three benchmark datasets, and present real-world robotic manipulation results.

\subsection{Experimental setup}
\label{sec:experimentalsetup}
\textbf{3D reconstruction models.} We evaluate SCOUT using four Image-to-3D models: Hunyuan3D~\cite{hunyuan3d22025tencent}, InstantMesh~\cite{xu2024instantmeshefficient3dmesh}, TripoSR~\cite{tochilkin2024triposrfast3dobject}, and TRELLIS~\cite{xiang2024structuredtrellis}. These models were selected to ensure architectural diversity, giving rise to complementary biases across object categories and input viewpoints. In addition, we include a view-invariant reconstruction option, approximated as a fixed score since our datasets contain objects with diffuse textures~\cite{downs2022google}, as discussed in Sec.~\ref{sec:problem_formulation}.

\textbf{Datasets and reconstruction pipeline.} Training the router requires a dataset of paired inputs and reconstruction scores. We begin with a collection of ground-truth 3D meshes and corresponding images, obtained either through rendering or real-world capture. For each dataset image, we run all models and score each reconstruction against the ground-truth mesh to produce $\mathbf{s}^{(i)} \in \mathbb{R}^k$.

We evaluate SCOUT on standard datasets: the Google Scanned Objects (GSO) dataset~\cite{downs2022google} and the YCB dataset~\cite{calli2015ycb} (supplemented with real images from BigBIRD~\cite{singh2014bigbird}). For the GSO dataset, which contains 1{,}030 objects, we render 15 images per object and 15{,}450 images in total. All viewpoints are nondegenerate, as the azimuth and elevation angles are chosen to avoid direct face-on views of the object.

For the YCB dataset, we select 54 objects that have corresponding real images from the BigBIRD dataset. For each object, we select 13 images spanning 5 elevation angles spaced from $0^\circ$ to $90^\circ$; the top view ($90^\circ$) is captured at a single azimuth angle, while the remaining elevations are captured at azimuth angles of $0^\circ$, $45^\circ$, and $90^\circ$. Unlike the GSO dataset, many of these viewpoints are degenerate, capturing the object face-on with limited depth information, making this a more challenging evaluation setting. For each viewpoint, we collect a real image from the BigBIRD dataset, a flash-style lighting render, and a surround-style lighting render. This yields 39 images per object and 2{,}106 images in total. The combination of challenging viewpoints, lighting variation, and limited data makes this setting substantially different from the GSO dataset.

For each image, we reconstruct a mesh with each model, normalize it to a unit sphere, register via FoundationPose~\cite{wen2024foundationpose} followed by DDM~\cite{ren2024ddm}, and then evaluate using a reconstruction quality metric. Our primary metric is Density-aware Chamfer Distance (DCD)~\cite{wu2021density}, which we negate so that higher scores indicate better reconstructions, with additional results reported for Chamfer Distance, IoU, and geometric metrics from Eval3D~\cite{duggal2025eval3d}.

\textbf{Baselines.} To the best of our knowledge, no prior routing methods exist for the 3D reconstruction domain; therefore, we adapt and evaluate several routing architectures from the LLM literature. Specifically, we compare against: (i) ZOOTER~\cite{lu2024routing}, which converts the full score vector---including view-invariant scores---to a probability distribution and trains a supervised router on the resulting targets; (ii) RouterDC~\cite{chen2024routerdc}, which learns a latent embedding space for each model as well as a query-dependent image representation using contrastive losses; (iii) an MLP baseline from RouterBench~\cite{hu2024routerbench}; (iv) a matrix factorization (MF) method based on RouteLLM~\cite{ong2024routellm}; and (v) kNN and linear regression (LR) models, which have been shown to frequently outperform more complex neural architectures in the LLM routing setting~\cite{li2025rethinking}. The most similar of these to our method is ZOOTER, with two key differences: SCOUT decouples viewpoint-dependent and view-invariant scores, and predicts scores directly rather than selecting the highest-ranked model, enabling support for arbitrary cost coefficient vectors and additional view-invariant methods without retraining.

We report performance relative to an input-agnostic baseline, which computes the average reconstruction score for each model across all training examples, yielding a fixed score estimate $\bar{s}(m_j) = \frac{1}{n}\sum_{i=1}^{n} s(x^{(i)}, m_j)$ per model, and selects $\arg\max_{m_j} \bar{s}(m_j) - \mathbf{c}[j]$.

For the evaluation metrics prescribed by RouterBench, we additionally include the Zero router baseline~\cite{hu2024routerbench}, which randomly samples routers at a given total cost allocation to trace out the non-decreasing convex hull (shown in \cref{fig:deferral_curves}). The Zero router is not included in other tables, as those evaluate average regret over cost coefficient vector subspaces rather than utility at a given total cost allocation.

\textbf{Image representation.} The router takes as input concatenated features from CLIP ViT-B/32~\cite{radford2021learningclip}, ResNet-50~\cite{he2016deep}, ViT-B/16~\cite{dosovitskiy2020image}, and ConvNeXt-B~\cite{liu2022convnet}.

\textbf{Evaluations.} Since reconstruction quality is continuous and metric-dependent, with no notion of a perfectly correct output, we evaluate routing performance in terms of average utility or average regret over various cost subspaces.

Similar to splitting $\mathbf{s}^{(i)}$ (Section~\ref{sec:scout}), we split the cost coefficient vector as $\mathbf{c} \equiv \left[\mathbf{c}_{\mathrm{dep}}, \mathbf{c}_{\mathrm{inv}}\right]$ for ease of notation. The first subspace is the single cost coefficient vector $\{\mathbf{c}_{\mathrm{dep}}=\mathbf{0}  \cap \mathbf{c}_{\mathrm{inv}}=\infty\} \equiv \mathcal{C}_0$, forcing selection among viewpoint-dependent models only. This also enables comparison with ZOOTER and RouterDC, which do not support varying costs at inference time. The second subspace consists of the one-dimensional cost families $\lambda \cdot \mathbf{c}_{\mathrm{latency}}$, $\lambda \cdot \mathbf{c}_{\mathrm{memory}}$, and $\lambda \cdot (\mathbf{c}_{\mathrm{latency}} \odot \mathbf{c}_{\mathrm{memory}})$, which penalize inference latency, memory, and memory-time occupancy (GB$\cdot$s), respectively, each swept over a range of $\lambda$ values. For these cost families, we report additional metrics including the cost-deferral curve and Average Improvement in Quality (AIQ)~\cite{hu2024routerbench}. As mesh reconstruction scores are continuous rather than discrete, we normalize the $y$-axis by the maximum achievable utility under an oracle router with no cost penalization, rather than reporting accuracy.

The third subspace is the full ``interesting'' cost subspace $\mathcal{C}$, defined to capture cost coefficient vectors that place both the viewpoint-dependent and view-invariant models in comparable utility ranges, thereby ensuring non-trivial routing decisions. For each method $m_j$, we compute the 75th percentile $P_{75,j}$ and interquartile range $\mathrm{IQR}_j$ of its score distribution. We then set the cost range for $c_j$ to $[P_{75,j} - \min_j(\mathrm{IQR}_j),\ P_{75,j}]$, where $\min_j(\mathrm{IQR}_j)$ ensures a uniform range width across all models.

\subsection{GSO results}
\Cref{tab:baseline_comparisons} shows the regret of each method across different cost subspaces on novel objects using DCD. SCOUT achieves statistically significant lower regret than all baselines across all cost subspaces. For subspace $\mathcal{C}_0$, ZOOTER performs most similarly to SCOUT. The benefit of decoupling becomes more pronounced in the broader subspace $\mathcal{C}$, where routing decisions are more sensitive to precise score predictions, as the routing decisions are all nontrivial.

\begin{table}[h]
    \centering
    \caption{Regret ($\downarrow$) as a proportion of the input-agnostic baseline regret across different cost subspaces on the GSO dataset.}
    \label{tab:baseline_comparisons}
    \resizebox{\columnwidth}{!}{
    \begin{tabular}{|l|c|c|c|}
    \hline
    \textbf{Method} & \textbf{$\{\mathbf{c} \in \mathcal{C}_0\}$} & \textbf{$\{\mathbf{c} \in \mathcal{C}\}$} & \textbf{$\{\mathbf{c} \in \mathcal{C}\cap\, \mathbf{c}_{\mathrm{inv}}=\infty\}$} \\
    \hline
    ZOOTER & $0.6489\pm0.0038$ & N/A & N/A \\
    \hline
    RouterDC & $0.8013\pm0.0050$ & N/A & N/A\\
    \hline
    MLP & $0.6716\pm0.0038$ & $0.4653\pm0.0026$ & $0.5004\pm0.0021$ \\
    \hline
    MF & $0.7557\pm0.0066$ & $0.5163\pm0.0036$    & $0.5814\pm0.0040$ \\
    \hline
    kNN & $0.6489\pm0.0033$ & $0.4630\pm0.0022$ & $0.4992\pm0.0019$ \\
    \hline
    LR & $0.6497\pm0.0033$ & $0.4519\pm0.0021$ & $0.5047\pm0.0019$ \\
    \hline
    Ours (no decoupling) & $0.6489\pm0.0038$ & $0.4511\pm0.0021$ & $0.4882\pm0.0018$ \\
    \hline
    Ours & $\mathbf{0.6386\pm0.0033}$ & $\mathbf{0.4319\pm0.0021}$ & $\mathbf{0.4831\pm0.0019}$ \\
    \hline
    Input-agnostic & $1.0000$ & $1.0000$ & $1.0000$ \\
    \hline
    Always HY3D & $3.0730\pm0.0091$ & $1.9693\pm0.0054$ & $1.2863\pm0.0063$ \\
    \hline
    Always IM & $1.9272\pm0.0049$ & $2.5788\pm0.0052$ & $1.8690\pm0.0040$ \\
    \hline
    Always TRL & $1.0000$ & $2.4762\pm0.0056$ & $1.7709\pm0.0053$ \\
    \hline
    Always TSR & $3.1118\pm0.0079$ & $2.5109\pm0.0057$ & $1.8041\pm0.0051$ \\
    \hline
    \end{tabular}
    }
\end{table}

\begin{figure}[h]
    \centering
    \begin{subfigure}[b]{0.49\linewidth}
        \centering
        \includegraphics[width=\linewidth]{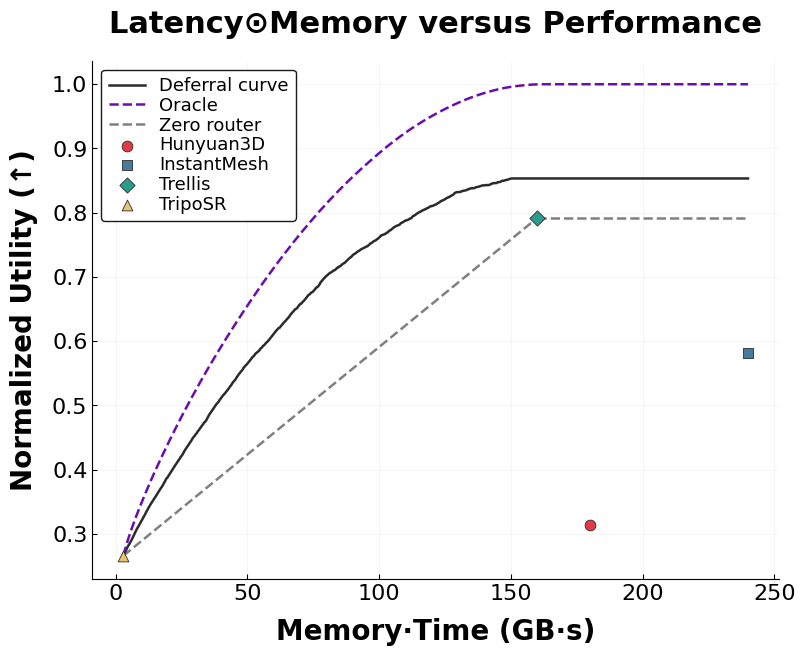}
        \caption{Latency$\odot$Memory deferral curve.}
        \label{fig:mem_deferral}
    \end{subfigure}
    \hfill
    \begin{subfigure}[b]{0.49\linewidth}
        \centering
        \includegraphics[width=\linewidth]{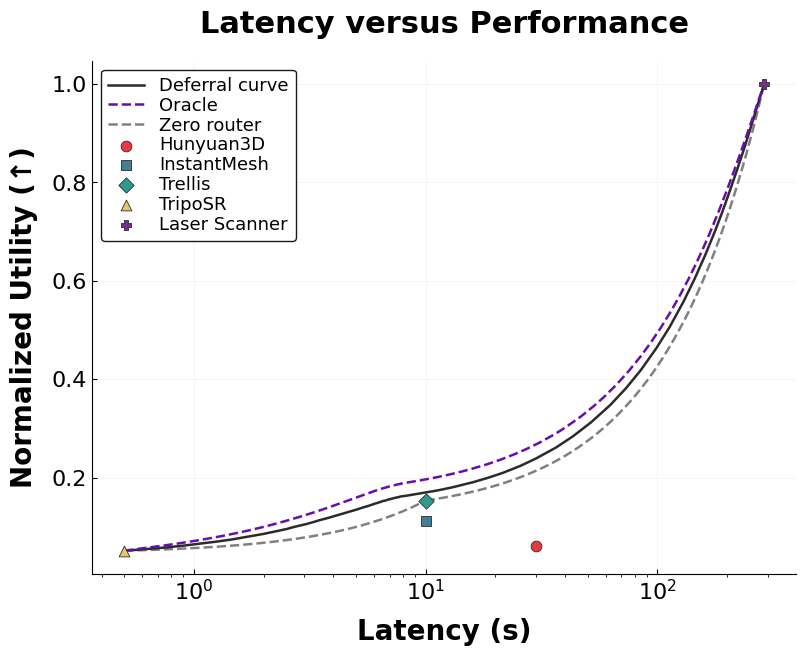}
        \caption{Latency deferral curve.\\$\space$}
        \label{fig:latency_deferral}
    \end{subfigure}
    \caption{Deferral curves for the latency$\odot$memory (a) and latency (b) cost coefficient vectors, showing the normalized utility achieved by each method as a function of the allocated cost.}
    \label{fig:deferral_curves}
    \vspace{-14px}
\end{figure}

For the one-dimensional cost families, we evaluate using the standard metrics from RouterBench. By sweeping $\lambda$, we obtain different achieved utilities at different total cost allocations, from which we can construct a Pareto frontier $\left(R_h(c)\right)$ as shown in \cref{fig:deferral_curves}. The AIQ, defined as:
\begin{equation}
    \mathrm{AIQ}(h) = \frac{1}{c_{max} - c_{min}} \int_{c_{min}}^{c_{max}} R_h\, \mathrm{d}c
\end{equation}
is shown in \cref{tab:AIQ}. ZOOTER and RouterDC are excluded from this evaluation as they do not support varying $\lambda$ without retraining. The deferral curves show that the presence of a high-quality but high-latency method (such as structured light scanning) reduces the margin for possible improvement, as the difference in AIQ between the Zero router and an oracle router is very small. SCOUT achieves the highest AIQ for memory and latency$\odot$memory, and is competitive on latency, where MLP performs equally well. The magnitude of improvement over baselines is consistent with the inter-method differences reported in RouterBench~\cite{hu2024routerbench}, where gains between routing methods are similarly modest.

\begin{table}[h]
    \centering
    \caption{AIQ ($\uparrow$) for one-dimensional cost families on the GSO dataset, evaluated on novel objects.}
    \resizebox{\columnwidth}{!}{
    \begin{tabular}{|l|c|c|c|}
    \hline
    \textbf{Method} & \textbf{Latency$\odot$Memory} & \textbf{Latency} & \textbf{Memory}\\
    \hline
    MLP & $0.7365\pm0.0008$ & $\mathbf{0.5815\pm0.0003}$ & $0.7911\pm0.0007$\\
    \hline
    MF &  $0.7233\pm0.0011$ & $0.5795\pm0.0003$ & $0.7736\pm0.0011$\\
    \hline
    kNN & $0.7382\pm0.0007$ & $0.5798\pm0.0002$ & $0.7942\pm0.0006$\\
    \hline
    LR &  $0.7364\pm0.0007$ & $0.5787\pm0.0002$ & $0.7913\pm0.0006$\\
    \hline
    Ours (no dc) & $0.7408\pm0.0007$ & $0.5796\pm0.0003$ & $0.7958\pm0.0006$\\
    \hline
    Ours & $\mathbf{0.7416\pm0.0007}$ & $0.5814\pm0.0002$ & $\mathbf{0.7976\pm0.0006}$\\
    \hline
   
    Zero router & $0.6248\pm0.0006$ & $0.5628\pm0.0002$ & $0.6506\pm0.0009$\\
    \hline
    Oracle & $0.8557\pm0.0006$ & $0.6066\pm0.0002$ & $0.9342\pm0.0003$\\
    \hline
    \end{tabular}
    }
    \label{tab:AIQ}
    \vspace{-1.5em}
\end{table}

\subsection{BigBIRD and YCB results}
We now evaluate on BigBIRD + YCB across multiple scoring metrics. Varying the metric is important because different metrics induce different optimal routing policies: IoU is a volume-based metric that strongly penalizes models that fill in hollow geometry, whereas DCD measures surface-level point correspondences and penalizes local geometric inaccuracies regardless of volume. Since different robotics tasks may require different quality criteria, a routing framework must perform well across metrics. We report results for $\mathcal{C}_0$ in \cref{tab:singlespace} and $\mathcal{C}$ in \cref{tab:fullspace} for evaluation on novel objects. On average, models produce the highest regret on Eval3D metrics, as these scores rely on generative models~\cite{liu2023zero1to3, yang2024depth}, introducing additional noise to an already challenging dataset. SCOUT is robust across metrics, from easier ones such as IoU (where all methods achieve relatively low regret) to noisier Eval3D metrics: it most frequently achieves the lowest regret and is the only method ranking in the top three across all metrics in \cref{tab:singlespace,tab:fullspace}.

\begin{table*}[h]
    \centering
    \caption{Regret ($\downarrow$) as a proportion of the input-agnostic baseline regret for $\mathcal{C}_0$ across multiple quality metrics on BigBIRD + YCB. Ours (no dc) denotes our method without decoupling, equivalent to ZOOTER extended to support arbitrary cost vectors.}
    \resizebox{\textwidth}{!}{
    \begin{tabular}{|l|c|c|c|c|c|c|c|}
    \hline
    \textbf{Method} & \textbf{DCD} & \textbf{Chamfer L2} & \textbf{Chamfer L1} & \textbf{IoU} & \textbf{MMD-EMD} & \textbf{Eval3D-geo} & \textbf{Eval3D-struct} \\
    \hline
    MLP & $1.1628\pm0.0147$ & $1.1677\pm0.0252$ & $1.0268\pm0.0156$ & $0.7135\pm0.0100$ & $0.9354\pm0.0147$ & $1.1520\pm0.0431$ & $1.2257\pm0.0209$ \\
    \hline
    MF &  $1.3116\pm0.0152$ & $1.3869\pm0.0247$ & $1.2151\pm0.0169$ & $0.8972\pm0.0172$ & $1.3003\pm0.0222$ & $1.6351\pm0.0791$ & $1.3365\pm0.0218$ \\
    \hline
    kNN & $1.0385\pm0.0118$ & $0.9109\pm0.0134$ & $0.8818\pm0.0093$ & $0.7411\pm0.0089$ & $0.9604\pm0.0158$ & $0.9957\pm0.0531$ & $1.1563\pm0.0158$ \\
    \hline
    LR &  $0.9983\pm0.0120$ & $0.9174\pm0.0147$ & $0.8506\pm0.0105$ & $\mathbf{0.6421\pm0.0090}$ & $0.8554\pm0.0139$ & $\mathbf{0.8246\pm0.0274}$ & $\mathbf{1.0815\pm0.0159}$ \\
    \hline
    Ours (no dc) & $1.0171\pm0.0129$ & $0.8806\pm0.0152$ & $0.8395\pm0.0121$ & $0.6849\pm0.0088$ & $0.8447\pm0.0143$ & $0.8674\pm0.0412$ & $1.1197\pm0.0167$\\
    \hline
    Ours & $\mathbf{0.9767\pm0.0113}$ & $\mathbf{0.8730\pm0.0146}$ & $\mathbf{0.8222\pm0.0106}$ & $0.6609\pm0.0091$ & $\mathbf{0.8136\pm0.0144}$ & $0.8588\pm0.0413$ & $1.1527\pm0.0188$\\
    \hline
    Input-agnostic & $1.0000$ & $1.0000$ & $1.0000$ & $1.0000$ & $1.0000$ & $1.0000$ & $1.0000$\\
    \hline
    \end{tabular}
    }
    \label{tab:singlespace}
    \vspace{-.5em}
\end{table*}

\begin{table*}[h]
    \centering
    \caption{Regret ($\downarrow$) as a proportion of the input-agnostic baseline regret for $\mathbf{c} \in \mathcal{C}$ across multiple quality metrics on BigBIRD + YCB. Ours (no dc) denotes our method without decoupling.}
    \resizebox{\textwidth}{!}{
    \begin{tabular}{|l|c|c|c|c|c|c|c|}
    \hline
    \textbf{Method} & \textbf{DCD} & \textbf{Chamfer L2} & \textbf{Chamfer L1} & \textbf{IoU} & \textbf{MMD-EMD} & \textbf{Eval3D-geo} & \textbf{Eval3D-struct} \\
    \hline
    MLP & $0.9944\pm0.0165$ & $1.1708\pm0.0234$ & $0.8983\pm0.0149$ & $0.7054\pm0.0115$ & $0.8153\pm0.0128$ & $3.1921\pm0.1267$ & $0.8975\pm0.0117$ \\
    \hline
    MF &  $1.2518\pm0.0158$ & $2.7886\pm0.0524$ & $1.7604\pm0.0281$ & $0.8378\pm0.0119$ & $1.8527\pm0.0301$ & $5.9481\pm0.2481$ & $1.3899\pm0.0197$ \\
    \hline
    kNN & $0.9845\pm0.0187$ & $0.7939\pm0.0096$ & $0.8121\pm0.0125$ & $0.7951\pm0.0121$ & $0.7699\pm0.0157$ & $\mathbf{0.8413\pm0.0122}$ & $0.7794\pm0.0094$ \\
    \hline
    LR &  $0.8925\pm0.0170$ & $0.8659\pm0.0090$ & $0.7776\pm0.0116$ & $0.7208\pm0.0099$ & $0.7080\pm0.0128$ & $1.0219\pm0.0161$ & $0.7366\pm0.0095$ \\
    \hline
    Ours (no dc) & $0.9451\pm0.0164$ & $\mathbf{0.7878\pm0.0095}$ & $0.7450\pm0.0109$ & $\mathbf{0.6767\pm0.0094}$ & $0.7147\pm0.0119$ & $0.9922\pm0.0227$ & $\mathbf{0.7336\pm0.0094}$ \\
    \hline
    Ours & $\mathbf{0.8880\pm0.0160}$ & $0.7991\pm0.0107$ & $\mathbf{0.7433\pm0.0105}$ & $0.6808\pm0.0101$ & $\mathbf{0.6741\pm0.0107}$ & $0.9253\pm0.0166$ & $0.7564\pm0.0136$ \\
    \hline
    Input-agnostic & $1.0000$ & $1.0000$ & $1.0000$ & $1.0000$ & $1.0000$ & $1.0000$ & $1.0000$\\
    \hline
    \end{tabular}
    }
    \label{tab:fullspace}
    \vspace{-.5em}
\end{table*}

\subsection{Robotics results}
We evaluate whether SCOUT selects reconstruction models that improve downstream robotic manipulation. We test across three tasks: collision-free grasp proposal generation in simulation, dexterous manipulation in simulation, and real-world pick-and-place on a Franka Panda robot.

\textbf{Grasp proposal evaluation.} Following~\cite{scenecomplete}, we generate grasp proposals on each reconstructed mesh using a Robotiq two-fingered gripper and evaluate against the ground truth. We report two metrics: the grasp collision rate (fraction of proposals that result in collision with the ground-truth geometry; lower is better) and mesh-IoU between the reconstruction and ground truth (higher is better). Results across 10 YCB objects are shown in \cref{tab:gc}. We evaluate SCOUT under two settings: routing over novel views of known objects, and the harder problem of routing over both novel views and novel objects. SCOUT achieves the lowest mean collision rate and highest mean mesh-IoU under both evaluation settings. The per-object results confirm that no single reconstruction model dominates across objects, and that per-input model selection translates to measurable gains in downstream grasp quality.

\begin{table*}[h]
    \centering
    \caption{Grasp collision rate ($\downarrow$) / mesh-IoU ($\uparrow$) for each reconstruction model across 10 YCB objects. Grasps are generated on the reconstructed mesh using a two-fingered gripper~\cite{da2_grasping} and evaluated against the ground-truth mesh. SCOUT is evaluated in two settings: routing over novel views of known objects, and routing over both novel views and novel objects.}
    \resizebox{\textwidth}{!}{
    \begin{tabular}{|l|c|c|c|c||c|c|}
    \hline
    \textbf{Object} & \textbf{Hunyuan3D} & \textbf{InstantMesh} & \textbf{TRELLIS} & \textbf{TripoSR} & \textbf{Ours (novel obj.)} & \textbf{Ours (novel view)} \\
    \hline
    Tomato Soup Can & $0.7174$ / $0.0685$ & $0.0227$ / $0.9164$ & $0.3864$ / $0.2071$ & $0.6585$ / $0.1807$ & InstantMesh & InstantMesh \\
    \hline
    Cup (vC)        & $0.0$ / $0.4019$   & $0.2273$ / $0.1381$ & $0.0$ / $0.4372$   & $0.1500$ / $0.1487$ & Hunyuan3D   & Hunyuan3D \\
    \hline
    Marble          & $0.0$ / $0.9580$   & $0.0$ / $0.7423$    & $0.0$ / $0.1464$   & $0.0$ / $0.9537$    & TripoSR   & TripoSR \\
    \hline
    Chef Can        & $0.0$ / $0.8543$   & $0.0$ / $0.8763$    & $0.0250$ / $0.9406$ & $0.1190$ / $0.7730$ & InstantMesh   & TRELLIS \\
    \hline
    Potted Meat     & $0.0$ / $0.6589$   & $0.6098$ / $0.4604$ & $0.0682$ / $0.7573$ & $0.3571$ / $0.4632$ & InstantMesh & TRELLIS \\
    \hline
    Cup (vB)        & $0.0$ / $0.2960$   & $0.0238$ / $0.1423$ & $0.1087$ / $0.0338$ & $0.3659$ / $0.1080$ & Hunyuan3D   & Hunyuan3D \\
    \hline
    Tennis Ball     & $0.0$ / $0.9717$   & $0.0$ / $0.9663$    & $0.0$ / $0.1525$   & $0.0$ / $0.9645$    & InstantMesh   & InstantMesh \\
    \hline
    Brick           & $0.0652$ / $0.2421$ & $0.0652$ / $0.2547$ & $0.6136$ / $0.0949$ & $0.0$ / $0.8334$   & TripoSR   & TripoSR \\
    \hline
    Power Drill     & $0.0714$ / $0.6230$ & $0.0500$ / $0.5662$ & $0.2889$ / $0.1399$ & $0.3636$ / $0.4793$ & InstantMesh & InstantMesh \\
    \hline
    Spatula         & $0.0851$ / $0.2218$ & $0.0465$ / $0.0076$ & $0.0851$ / $0.1014$ & $0.0$ / $0.0392$   & InstantMesh & Hunyuan3D \\
    \hline\hline
    Mean            & $0.0939$ / $0.5296$ & $0.1045$ / $0.5071$ & $0.1576$ / $0.3011$ & $0.2014$ / $0.4944$ & $0.0729$ / $0.6278$ & $0.0251$ / $0.6854$ \\
    \hline
    \end{tabular}
    }
    \label{tab:gc}
    \vspace{-.5em}
\end{table*}

\textbf{Dexterous manipulation.} We next evaluate whether reconstruction quality affects dexterous grasping success. Following~\cite{scenecomplete}, we test each reconstruction model in simulation across 5 objects and report the dexterous grasp success rate. Results are shown in \cref{tab:dexterous_results}. Dexterous manipulation places stricter demands on surface accuracy than two-fingered grasping, as contact placement and force closure depend on local surface geometry. This setting thus amplifies the performance differences between reconstruction models and further motivates per-input model selection.

\textbf{Real-world pick and place.} Finally, we validate the routing framework on a physical Franka Panda robot performing tabletop pick-and-place. For each of 5 objects, we reconstruct the object from a single input image using each Image-to-3D method, generate antipodal grasp proposals using DA2~\cite{da2_grasping}, and randomly sample 5 proposals for execution. We report the pick-and-place success rate in \cref{tab:realworld_results}. This experiment tests the full pipeline end-to-end: a single input image is routed to a reconstruction model, the resulting mesh is used for grasp planning, and the planned grasp is executed on the robot. The router's selections reflect the difficulty of the input viewpoint: for objects 1, 3, and 4, where the viewpoints are more challenging, SCOUT selects Hunyuan3D, which produces the highest-quality reconstructions under difficult conditions. For objects 2 and 5, where the viewpoints are more straightforward, the router selects TRELLIS, which suffices for these easier inputs and has lower latency.

\begin{table*}[h]
  \begin{minipage}{0.48\textwidth}
    \centering
    \caption{Dexterous manipulation success rate ($\uparrow$) in simulation for each reconstruction model across 5 objects (HY3D: Hunyuan3D, IM: InstantMesh, TRL: TRELLIS, TSR: TripoSR). Ours reports the model selected by SCOUT.}
    \resizebox{\columnwidth}{!}{
    \begin{tabular}{|l|c|c|c|c|c|c|}
    \hline
    \textbf{Method} & \textbf{Can} & \textbf{Spatula} & \textbf{Drill} & \textbf{Brick} & \textbf{Cup} & \textbf{Mean} \\
    \hline
    HY3D         & $13.6$ & $15.2$ & $38.1$ & $61.8$ & $49.4$ & $35.6$ \\
    \hline
    IM           & $29.9$ & $0$ & $41.7$ & $52.9$ & $44.5$ & $33.8$ \\
    \hline
    TRL          & $61.8$ & $0$ & $4.5$ & $50.4$ & $45.8$ & $32.5$ \\
    \hline
    TSR          & $21.9$ & $0$ & $17.6$ & $39.7$ & $45.0$ & $24.8$ \\
    \hline\hline
    \textbf{Ours} & TRL & HY3D & IM & TSR & HY3D & $41.6$ \\
    \hline
    \end{tabular}
    }
    \label{tab:dexterous_results}
  \end{minipage}
  \hfill
  \begin{minipage}{0.48\textwidth}
    \centering
    \caption{Real-world pick-and-place success rate ($\uparrow$) on a Franka Panda robot across 5 objects. For each reconstructed object, 5 antipodal grasps are sampled and executed.}
    \resizebox{\columnwidth}{!}{
    \begin{tabular}{|l|c|c|c|c|c|c|}
    \hline
    \textbf{Method} & \textbf{Drill} & \textbf{Ball} & \textbf{Can} & \textbf{Cup} & \textbf{Mug} & \textbf{Mean} \\
    \hline
    HY3D         & $3/5$ & $4/5$ & $5/5$ & $5/5$ & $5/5$ & $4.4/5$ \\
    \hline
    IM           & $1/5$ & $5/5$ & $1/5$ & $2/5$ & $4/5$ & $2.6/5$ \\
    \hline
    TRL          & $0/5$ & $5/5$ & $3/5$ & $0/5$ & $5/5$ & $2.6/5$ \\
    \hline
    TSR          & $1/5$ & $0/5$ & $2/5$ & $0/5$ & $0/5$ & $0.6/5$ \\
    \hline\hline
    \textbf{Ours} & HY3D & TRL & HY3D & HY3D & TRL & $4.6/5$ \\
    \hline
    \end{tabular}
    }
    \label{tab:realworld_results}
  \end{minipage}
\end{table*}
\section{Conclusion}
\label{sec:conclusion}

We presented SCOUT, a routing framework that selects among 3D reconstruction models for a given input image under arbitrary cost-quality tradeoffs. The key idea is to decouple reconstruction scores into relative model performance and image difficulty, which keeps the learned network independent of the number of view-invariant methods and supports arbitrary cost coefficient vectors at inference time without retraining. SCOUT consistently outperforms routing baselines adapted from the LLM literature across three datasets, multiple quality metrics, and diverse cost coefficient vector subspaces, with the largest gains in the most challenging cost regions. Robotic manipulation experiments in simulation and the real world confirm routing improvements translate to downstream robotic performance.

\textit{Limitations \& Future Work}: Our evaluation uses four Image-to-3D (viewpoint-dependent) models; scaling to a larger model pool is a natural next step. Additionally, while our formulation supports view-invariant methods, our experiments approximate them with fixed scores rather than evaluating with real scanning hardware; validating with actual view-invariant pipelines is an important direction for future work. Integrating the router into a closed-loop manipulation system, where task feedback informs future routing decisions, is another promising direction.
\clearpage
\newpage
\clearpage
\section{Acknowledgments}
\label{sec:acknowledgments}
We gratefully acknowledge support from NSF grant 2214177; from AFOSR grant FA9550-22-1-0249; from ONR MURI grants N00014-22-1-2740 and N00014-24-1-2603; from the MIT Siegel Family Quest for Intelligence; and from the Robotics and AI Institute.
{
    \small
    \bibliographystyle{ieeenat_fullname}
    \bibliography{main}
}

\clearpage
\maketitlesupplementary

\section{Derivation of optimal weights}
\label{sec:derivation}

We derive global inverse-variance weights for combining $k_1$ estimators of the partition function. Each method $j\in\{1,\dots,k_1\}$ produces, for data point~$i$, the estimate
\begin{equation}\label{eq:ztilde}
    \tilde{z}_j^{(i)} = \frac{e^{s_j^{(i)}/T}}{\hat{p}_j^{(i)}}\,,
\end{equation}
and we seek weights $w_j$, independent of~$i$, for the combined estimate
\begin{equation}\label{eq:combo}
    \tilde{z}^{(i)} = \sum_{j=1}^{k_1} w_j\,\tilde{z}_j^{(i)}.
\end{equation}

\textbf{Assumptions.}
\begin{enumerate}
    \item[\textbf{A1.}] \textbf{(Error model).} For each method~$j$ and data point~$i$,
    $\hat{p}_j^{(i)} = p^*_j{}^{(i)} + \epsilon_j^{(i)}$,
    where $p^*_j{}^{(i)}$ is the true (fixed) probability and $\epsilon_j^{(i)}$ is random noise with
    $\mathbb{E}_\epsilon\!\big[\epsilon_j^{(i)}\big] = 0$ and
    $\operatorname{Var}_\epsilon\!\big[\epsilon_j^{(i)}\big] = \sigma_j^2$.
    The noise variance $\sigma_j^2$ is a property of method~$j$ alone and does not depend on~$i$.

    \item[\textbf{A2.}] \textbf{(Small noise).} $|\epsilon_j^{(i)}| \ll p^*_j{}^{(i)}$ for all~$j,i$.

    \item[\textbf{A3.}] \textbf{(Uncorrelated methods).} $\operatorname{Cov}_\epsilon\!\big[\epsilon_j^{(i)},\,\epsilon_k^{(i)}\big] = 0$ for $j\neq k$.

    \item[\textbf{A4.}] \textbf{(Consistency).} All methods estimate the same true quantity: $z^{(i)} \equiv e^{{s}_j^{(i)}/T}\big/p^*_j{}^{(i)}$ is independent of~$j$.

    \item[\textbf{A5.}] \textbf{(Noise--signal independence).} $\epsilon_j^{(i)}$ is independent  of $p^*_j{}^{(i)}$ across data points; \ie, the estimation error does not depend on the true probability.
\end{enumerate}

Here $\mathbb{E}_\epsilon$ and $\operatorname{Var}_\epsilon$ denote expectation and variance over estimation noise at fixed $i$; all other quantities are deterministic for a given $i$.

\textbf{Global weight optimization}. Fixing a data point $i$ and expanding \cref{eq:ztilde} to first order in $\epsilon_j^{(i)}/p^*_j{}^{(i)}$:
\begin{equation}\label{eq:taylor}
    \tilde{z}_j^{(i)}
    = \frac{e^{{s}_j^{(i)}/T}}{p^*_j{}^{(i)} + \epsilon_j^{(i)}}
    \;\approx\;
    {z^{(i)}}
    \;-\;
    {\frac{z^{(i)}}{p^*_j{}^{(i)}}}\;\epsilon_j^{(i)}.
\end{equation}
Hence
\begin{equation}\label{eq:varzj}
    \operatorname{Var}_\epsilon\!\big[\tilde{z}_j^{(i)}\big]
    = \left(\frac{z^{(i)}}{p^*_j{}^{(i)}}\right)^{\!2}\sigma_j^2.
\end{equation}
Substituting $p^*_j{}^{(i)} = e^{{s}_j^{(i)}/T}/z^{(i)}$ from~\textbf{A4}:
\begin{equation}\label{eq:tau}
    \tau_j^{2,(i)}
    \;\equiv\;
    \operatorname{Var}_\epsilon\!\big[\tilde{z}_j^{(i)}\big]
    \;=\;
    \frac{\big(z^{(i)}\big)^4}{e^{2{s}_j^{(i)}/T}}\cdot\sigma_j^2.
\end{equation}
Inserting \cref{eq:taylor} into \cref{eq:combo}:
\begin{equation}
    \tilde{z}^{(i)} - z^{(i)}
    \;\approx\; -\sum_j w_j\,\frac{z^{(i)}}{p^*_j{}^{(i)}}\,\epsilon_j^{(i)}.
\end{equation}
Taking the variance over the joint noise $\{\epsilon_j^{(i)}\}_j$ and using~\textbf{A3}:
\begin{equation}\label{eq:varcombo}
    \operatorname{Var}_\epsilon\!\big[\tilde{z}^{(i)}\big]
    = \sum_j w_j^2\,\tau_j^{2,(i)}.
\end{equation}
The quantity \cref{eq:varcombo} depends on~$i$ through $\tau_j^{2,(i)}$. Since we seek weights independent of~$i$, we minimize the variance \emph{averaged} over data points:
\begin{equation}\label{eq:obj}
\begin{aligned}
    J(\mathbf{w})
    &\;=\;
    \mathbb{E}_i\!\big[\operatorname{Var}_\epsilon\!\big[\tilde{z}^{(i)}\big]\big]\\
    &\;=\;
    \mathbb{E}_i\!\bigg[\sum_j w_j^2\,\tau_j^{2,(i)}\bigg]\\
    &\;=\;
    \sum_j w_j^2\;{\mathbb{E}_i\!\big[\tau_j^{2,(i)}\big]}\\
    &\;\equiv\;
    \sum_j w_j^2\;{\bar{\tau}_j^{\,2}}
\end{aligned}
\end{equation}
Substituting \cref{eq:tau}:
\begin{equation}\label{eq:taubar}
    \bar{\tau}_j^{\,2}
    \;=\;
    \sigma_j^2\;\mathbb{E}_i\!\!\left[\frac{\big(z^{(i)}\big)^4}{e^{2{s}_j^{(i)}/T}}\right],
\end{equation}
We minimize \cref{eq:obj} subject to $\sum_j w_j = 1$. The Lagrangian is
\begin{equation}
    \mathcal{L}
    = \sum_j w_j^2\;\bar{\tau}_j^{\,2}
    - \lambda\!\left(\sum_j w_j - 1\right).
\end{equation}
Setting $\partial\mathcal{L}/\partial w_j = 2\,w_j\,\bar{\tau}_j^{\,2} - \lambda = 0$ gives $w_j = \lambda/(2\bar{\tau}_j^{\,2})$.
Substituting into the constraint:
\begin{equation}
    \sum_j \frac{\lambda}{2\bar{\tau}_j^{\,2}} = 1
    \quad\Longrightarrow\quad
    \lambda = \frac{2}{\displaystyle\sum_k 1/\bar{\tau}_k^{\,2}}\,.
\end{equation}
The Hessian of $\mathcal{L}$ restricted to the constraint surface has diagonal entries $2\bar{\tau}_j^{\,2}>0$, confirming this is a minimum. The optimal global weights are
\begin{equation}\label{eq:global}
    \boxed{
    w_j
    \;=\;
    \frac{1/\bar{\tau}_j^{\,2}}{\displaystyle\sum_k 1/\bar{\tau}_k^{\,2}}
    \;\propto\;
    \frac{1}{\bar{\tau}_j^{\,2}}
    \;=\;
    \frac{1}{\;\sigma_j^2\;\mathbb{E}_i\!\!\left[\dfrac{(z^{(i)})^4}{e^{2{s}_j^{(i)}/T}}\right]}
    \,.
    }
\end{equation}

\textbf{Estimation of $\sigma^2_j$}. It remains to estimate the noise variance $\sigma_j^2$ from data. Define the following quantities computed across data points:
\begin{equation}
    S_j^2 \;\equiv\; \operatorname{Var}_i\!\big[\hat{p}_j^{(i)}\big], \qquad
    R_j^2 \;\equiv\; \operatorname{Corr}_i\!\big[p^*_j{}^{(i)},\;\hat{p}_j^{(i)}\big]^2.
\end{equation}

Using the error model $\hat{p}_j^{(i)} = p^*_j{}^{(i)} + \epsilon_j^{(i)}$, the independence assumption~\textbf{A5} gives $\operatorname{Cov}_i[p^*_j{},\,\epsilon_j] = 0$, and since $\sigma_j^2$ is constant across~$i$ (\textbf{A1}), we have:
\begin{equation}\label{eq:vardecomp}
    S_j^2 = \operatorname{Var}_i\!\big[p^*_j{}^{(i)} + \epsilon_j^{(i)}\big] = \operatorname{Var}_i\!\big[p^*_j{}^{(i)}\big] + \sigma_j^2.
\end{equation}
Using $\operatorname{Cov}_i[p^*_j{},\;\hat{p}_j] = \operatorname{Cov}_i[p^*_j{},\;p^*_j{}+\epsilon_j] = \operatorname{Var}_i[p^*_j{}]$, we find:
\begin{equation}\label{eq:r2}
    R_j^2
    = \frac{\operatorname{Cov}_i[p^*_j{},\;\hat{p}_j]^2}{\operatorname{Var}_i[p^*_j{}]\cdot S_j^2}
    = \frac{\operatorname{Var}_i[p^*_j{}]^2}{\operatorname{Var}_i[p^*_j{}]\cdot S_j^2}
    = \frac{\operatorname{Var}_i[p^*_j{}]}{S_j^2}\,.
\end{equation}
Substituting $\operatorname{Var}_i[p^*_j{}^{(i)}] = R_j^2\,S_j^2$ into \cref{eq:vardecomp}:
\begin{equation}\label{eq:sigma}
    \boxed{\sigma_j^2 \;=\; S_j^2\,(1 - R_j^2).}
\end{equation}

\section{Effect of smoothing on score recovery}
\label{sec:smoothing_approximation}

In Section~\ref{sec:scout}, we train $\hat{\mathbf{p}}$ on the softmax of the smoothed scores $\tilde{\mathbf{s}}_{\mathrm{dep}}^{(i)}$, but the recovered scores $\hat{s}(x^{(i)}, m_j) = T \cdot \ln\!\big(\hat{p}_j^{(i)} \cdot \hat{z}^{(i)}\big)$ should approximate the true unsmoothed scores $s_{\mathrm{dep},j}^{(i)}$ for comparison with $\mathbf{s}_{\mathrm{inv}}$. Here we show that the proxy construction targets these true scores by design. Throughout this section we write $s_j^{(i)} \equiv s_{\mathrm{dep},j}^{(i)}$ for brevity.

\textbf{Role of smoothing.}
The purpose of tag-based smoothing is to reduce noise in the training targets, producing more accurate predictions $\hat{\mathbf{p}}$ of relative model performance---the smoothed scores $\tilde{\mathbf{s}}_{\mathrm{dep}}^{(i)}$ are not themselves the quantities we wish to recover. The ablation in \cref{tab:scountsmoothing} confirms that smoothing improves routing performance, validating this role. Smoothing is not applied when constructing the proxy targets for $\hat{z}$, because $\hat{z}$ must capture the absolute difficulty of each individual input image, and smoothing suppresses precisely this per-image variation by blending scores toward category averages. This is consistent with the baseline ablation (\cref{tab:baselinessmoothing}): applying smoothing to LR and kNN, which predict absolute scores rather than relative probabilities, degrades their performance.

\begin{table}[h]
    \centering
    \caption{Regret ($\downarrow$) as a proportion of the input-agnostic baseline regret across different cost subspaces on the GSO dataset.}
    \label{tab:scountsmoothing}
    \resizebox{\columnwidth}{!}{
    \begin{tabular}{|l|c|c|c|}
    \hline
    \textbf{Method} & \textbf{$\{\mathbf{c} \in \mathcal{C}_0\}$} & \textbf{$\{\mathbf{c} \in \mathcal{C}\}$} & \textbf{$\{\mathbf{c} \in \mathcal{C}\cap\, \mathbf{c}_{\mathrm{inv}}=\infty\}$} \\
    \hline
    \makecell[l]{Ours w/\\smoothing} & $0.6386\pm0.0033$ & $0.4319\pm0.0021$ & $0.4831\pm0.0019$\\
    \hline
    \makecell[l]{Ours w/o\\smoothing} & $0.6588\pm0.0035$ & $0.4371\pm0.0021$ & $0.4878\pm0.0020$ \\
    \hline
    \end{tabular}
    }
    \vspace{0em}
\end{table}

\begin{table}[h]
    \centering
    \caption{Regret ($\downarrow$) as a proportion of the input-agnostic baseline regret across different cost subspaces on the GSO dataset.}
    \label{tab:baselinessmoothing}
    \resizebox{\columnwidth}{!}{
    \begin{tabular}{|l|c|c|c|}
    \hline
    \textbf{Method} & \textbf{$\{\mathbf{c} \in \mathcal{C}_0\}$} & \textbf{$\{\mathbf{c} \in \mathcal{C}\}$} & \textbf{$\{\mathbf{c} \in \mathcal{C}\cap\, \mathbf{c}_{\mathrm{inv}}=\infty\}$} \\
    \hline
    \makecell[l]{LR w/\\smoothing} & $0.6515\pm0.0034$ & $0.4723\pm0.0022$ & $0.5064\pm0.0019$\\
    \hline
    \makecell[l]{LR w/o\\smoothing} & $0.6497\pm0.0033$ & $0.4519\pm0.0021$ & $0.5047\pm0.0019$ \\
    \hline
    \makecell[l]{kNN w/\\smoothing} & $0.6515\pm0.0034$ & $0.4723\pm0.0022$ & $0.5064\pm0.0019$ \\
    \hline
    \makecell[l]{kNN w/o\\smoothing} & $0.6489\pm0.0033$ & $0.4630\pm0.0022$ & $0.4992\pm0.0019$ \\
    \hline
    \end{tabular}
    }
    \vspace{0em}
\end{table}

\textbf{Proxy construction targets true scores.}
Given $\hat{\mathbf{p}}$, we wish to find $\hat{z}^{(i)}$ such that $T \cdot \ln\!\big(\hat{p}_j^{(i)} \cdot \hat{z}^{(i)}\big) = s_j^{(i)}$. Solving for each model~$j$ individually gives the per-model proxy:
\begin{equation}
    \tilde{z}_j^{(i)} \;=\; \frac{e^{s_j^{(i)}/T}}{\hat{p}_j^{(i)}}, \label{eq:proxy_def}
\end{equation}
which uses the true unsmoothed score $s_j^{(i)}$ in the numerator. By construction:
\begin{equation}
    T \cdot \ln\!\big(\hat{p}_j^{(i)} \cdot \tilde{z}_j^{(i)}\big)
    \;=\; T \cdot \ln\!\left(\hat{p}_j^{(i)} \cdot \frac{e^{s_j^{(i)}/T}}{\hat{p}_j^{(i)}}\right)
    \;=\; s_j^{(i)}. \label{eq:exact_recovery}
\end{equation}
Each proxy exactly recovers the true score for model~$j$, regardless of what $\hat{p}_j^{(i)}$ was trained on, because the proxy is defined directly from the true scores.

\section{Implementation details}

\textbf{SCOUT implementation details.} The probability network $\hat{\mathbf{p}}$ is a feedforward neural network with two hidden layers of 128 units each, ReLU activations, and dropout ($p = 0.2$), trained with KL divergence loss using Adam (learning rate $10^{-4}$, weight decay $10^{-4}$) with batch size 128 for 10 epochs. Tag smoothing uses $\beta = 0.7$ and softmax temperature $T = 1.0$ for the GSO dataset. For BigBIRD + YCB, we decrease $\beta$ to $0.5$ because the scores are noisier, requiring stronger smoothing toward category averages. The softmax temperature $T$ is set per metric to account for differences in score scale (\cref{tab:temperatures}).

\begin{table}[h]
    \centering
    \caption{Softmax temperature $T$ per metric for BigBIRD + YCB.}
    \label{tab:temperatures}
    \resizebox{\columnwidth}{!}{
    \begin{tabular}{|l|c|c|c|c|c|c|c|}
    \hline
    \textbf{Metric} & \textbf{DCD} & \textbf{Chamfer L2} & \textbf{Chamfer L1} & \textbf{IoU} & \textbf{MMD-EMD} & \textbf{Eval3D-geo} & \textbf{Eval3D-struct} \\
    \hline
    $T$ & $0.5$ & $0.1$ & $0.2$ & $2.0$ & $0.15$ & $0.1$ & $0.2$ \\
    \hline
    \end{tabular}
    }
    \vspace{-1.5em}
\end{table}

The partition function $\hat{z}$ is predicted by Ridge regression ($\alpha = 10^3$), with the weighting terms $R_j^2$, $S_j^2$, and the expectation in $w_j$ estimated via 5-fold cross-validation over the training set. Image features are concatenated from CLIP ViT-B/32 (512-d) \cite{radford2021learningclip}, ResNet-50 (2048-d)
\cite{he2016deep}, ViT-B/16 (768-d) \cite{dosovitskiy2020image}, and ConvNeXt-B (1024-d) \cite{liu2022convnet}, yielding a 4352-dimensional input. Hyperparameters were selected on seeds 0-49. All final experiments were conducted on a single NVIDIA RTX 4090 GPU and averaged over seeds 50-199.

\begin{figure}[h]
    \centering
    \begin{subfigure}[b]{0.32\linewidth}
        \centering
        \includegraphics[width=\linewidth]{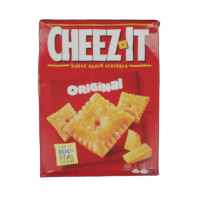}
        \caption{Degenerate view showing only the front face.}
    \end{subfigure}
    \hfill
    \begin{subfigure}[b]{0.32\linewidth}
        \centering
        \includegraphics[width=\linewidth]{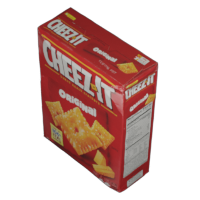}
        \caption{Nondegenerate view revealing full geometry.}
    \end{subfigure}
    \hfill
    \begin{subfigure}[b]{0.32\linewidth}
        \centering
        \includegraphics[width=\linewidth]{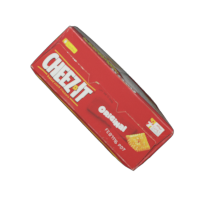}
        \caption{Degenerate top-down view.\\{}}
    \end{subfigure}
    \caption{Degenerate and nondegenerate views of a cracker box.}
    \label{fig:degenimage}
    \vspace{0px}
\end{figure}

\begin{figure}[h]
    \centering
    \begin{subfigure}[b]{0.32\linewidth}
        \centering
        \includegraphics[width=\linewidth]{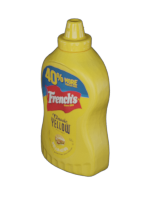}
        \caption{Rendered image with flash-style lighting.}
    \end{subfigure}
    \hfill
    \begin{subfigure}[b]{0.32\linewidth}
        \centering
        \includegraphics[width=\linewidth]{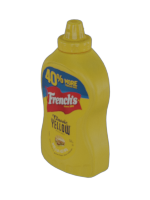}
        \caption{Rendered image with surround-style lighting.}
    \end{subfigure}
    \hfill
    \begin{subfigure}[b]{0.32\linewidth}
        \centering
        \includegraphics[width=\linewidth]{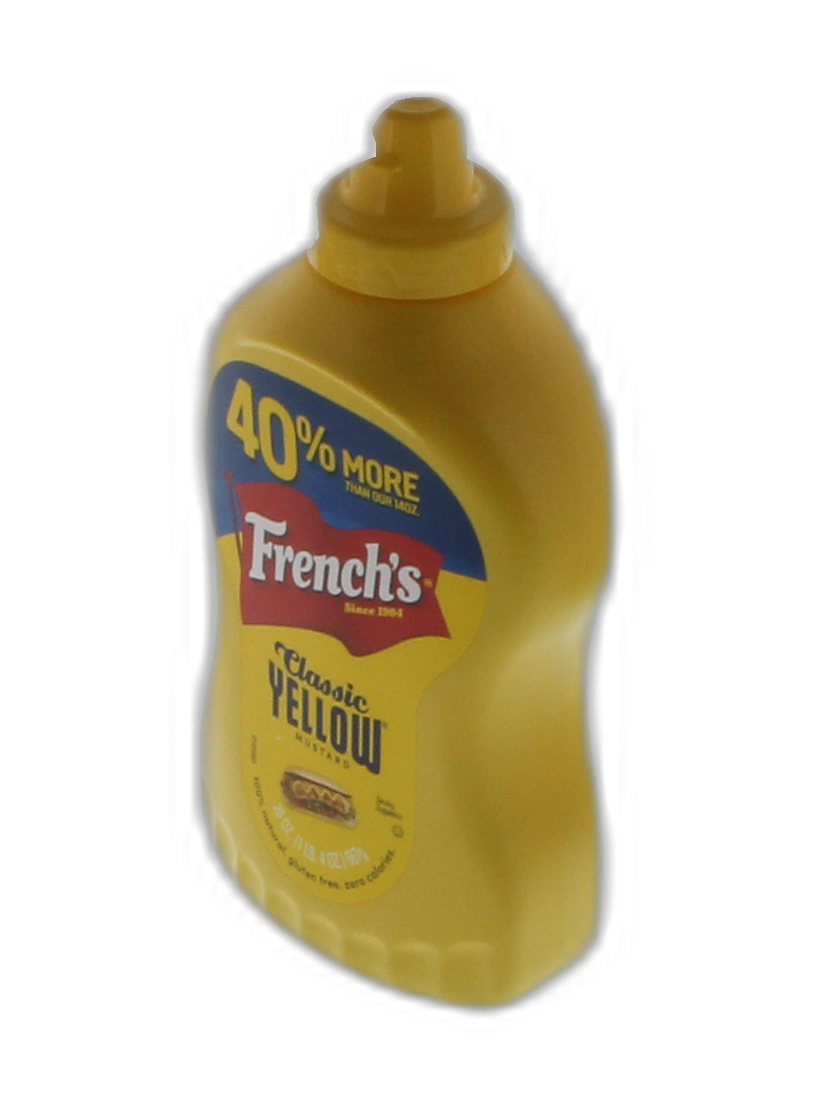}
        \caption{Real image from BigBIRD~\cite{singh2014bigbird}. \\{}}
    \end{subfigure}
    \caption{Different lighting conditions for a fixed viewpoint.}
    \label{fig:lightimage}
    \vspace{0px}
\end{figure}

\textbf{Data collection.} Scores were collected as follows. We began with ground-truth meshes from the GSO~\cite{downs2022google} and YCB~\cite{calli2015ycb} datasets, and collected images of each mesh at multiple viewpoints. For the GSO dataset, images were rendered at elevation angles of $35^\circ, 45^\circ, 60^\circ, 75^\circ, \text{ and } 85^\circ$. All elevations were captured at azimuth angles of $30^\circ, 120^\circ, \text{ and } 210^\circ$. This combination ensures all captured views are nondegenerate. For the YCB dataset, images were captured at elevation angles of $5^\circ, 22.5^\circ, 45^\circ, 67.5^\circ, \text{ and } 90^\circ$. The top view ($90^\circ$) was captured at a single azimuth angle of $45^\circ$, while all other elevations were captured at azimuth angles of $0^\circ, 45^\circ, \text{ and } 90^\circ$. Unlike GSO, this combination of elevation and azimuth angles produces many degenerate viewpoints. Each YCB viewpoint was captured under three conditions: flash-style lighting, surround-style lighting, and a real image from the BigBIRD dataset~\cite{singh2014bigbird}. Examples of degenerate versus nondegenerate views are shown in \cref{fig:degenimage} and examples of different lighting styles are shown in \cref{fig:lightimage}. Surround-style lighting is typically more challenging for Image-to-3D models than flash-style lighting, as the lack of shadows removes geometric cues that aid reconstruction. 


Once images were collected, we generated a reconstruction with each of the four Image-to-3D models: Hunyuan3D~\cite{hunyuan3d22025tencent} (version 2.0 throughout), InstantMesh~\cite{xu2024instantmeshefficient3dmesh}, TRELLIS~\cite{xiang2024structuredtrellis}, and TripoSR~\cite{tochilkin2024triposrfast3dobject}, using default parameters for all models. For Hunyuan3D, we applied a convex decomposition using CoACD~\cite{wei2022coacd} (concavity threshold 0.025, 50 MCTS search iterations, default remaining parameters) between the mesh generation and texture generation stages to prevent the texturing process from taking excessive time.

The ground-truth and reconstructed meshes were then normalized to a common scale by fitting each mesh to a unit sphere centered at the origin using \textit{miniball}~\cite{devert2023miniball}. We then registered each reconstructed mesh to its ground-truth counterpart via FoundationPose~\cite{wen2024foundationpose} for global registration (default parameters), followed by DDM~\cite{ren2024ddm} for local refinement (learning rate $2\text{e-}2$, 100 iterations).

\textbf{Cost vectors.} The AIQ and deferral curve metrics require latency, memory, and latency$\odot$memory cost vectors. Latency costs are 30s (Hunyuan3D), 10s (InstantMesh), 10s (TRELLIS), and 0.5s (TripoSR). Memory costs are 6GB (Hunyuan3D), 24GB (InstantMesh), 16GB (TRELLIS), and 6GB (TripoSR). All values reflect mesh generation only (excluding texturing) and are based on official documentation~\cite{hunyuan3d2repo, instantmeshrepo, trellisrepo, triposrrepo} when available, otherwise approximated by experiments.

\section{Ablations}

\begin{figure}[h]
    \centering
    \begin{subfigure}[b]{0.49\linewidth}
        \centering
        \includegraphics[width=\linewidth]{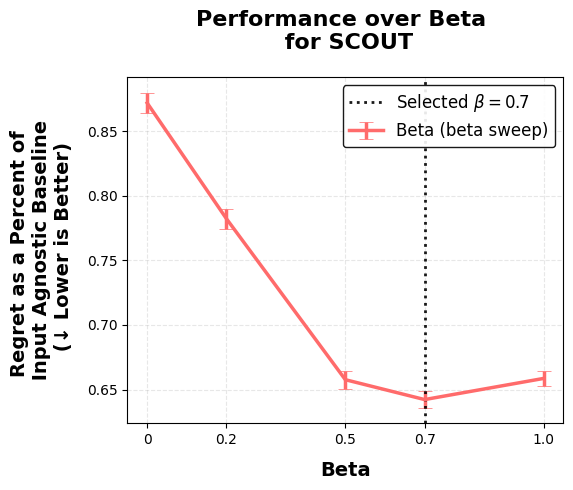}
        \caption{Beta sweep.}
    \end{subfigure}
    \hfill
    \begin{subfigure}[b]{0.49\linewidth}
        \centering
        \includegraphics[width=\linewidth]{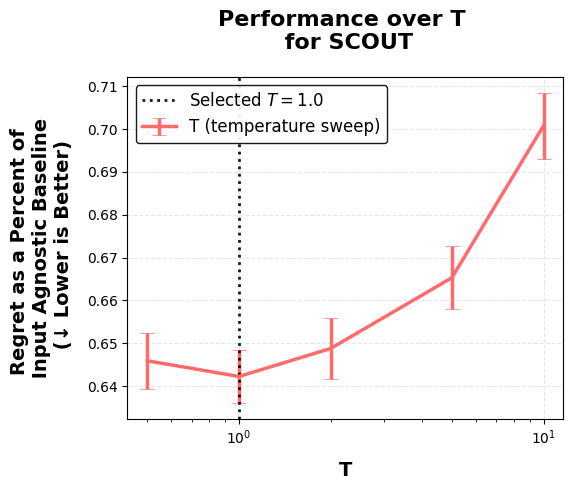}
        \caption{Temperature sweep.}
    \end{subfigure}

    \vspace{0.5em}

    \begin{subfigure}[b]{0.49\linewidth}
        \centering
        \includegraphics[width=\linewidth]{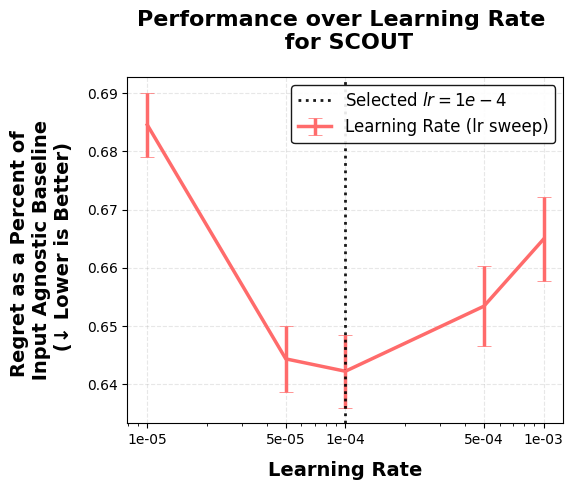}
        \caption{Learning rate sweep.}
    \end{subfigure}
    \hfill
    \begin{subfigure}[b]{0.49\linewidth}
        \centering
        \includegraphics[width=\linewidth]{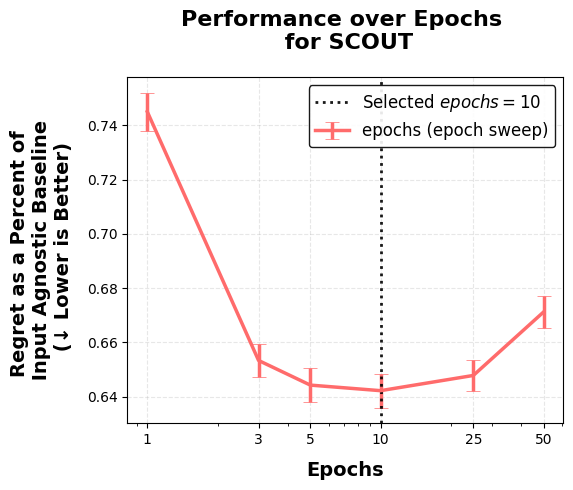}
        \caption{Epoch sweep.}
    \end{subfigure}
    \caption{Sensitivity of hyperparameters for SCOUT evaluated on novel objects in the GSO dataset over the cost coefficient vector $\mathcal{C}_0$.}
    \label{fig:scout_hyper}
    \vspace{0px}
\end{figure}

\begin{figure}
    \centering
    \includegraphics[width=\linewidth]{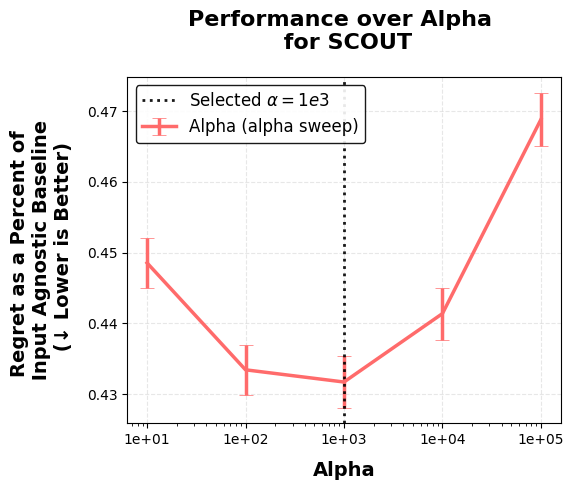}
    \caption{Sensitivity of alpha for SCOUT evaluated on novel objects in the GSO dataset over cost coefficient vectors sampled from $\mathcal{C}$.}
    \label{fig:scout_alpha}
    \vspace{0px}
\end{figure}

\textbf{SCOUT hyperparameters.} We perform one-at-a-time sweeps over each hyperparameter, holding the others fixed, to verify that the loss landscape is smooth with a local optimum near our selected values. These experiments were conducted on seeds 0-49. The network architecture (two hidden layers of 128 units each) was fixed to ensure fair comparison with the baselines. For the GSO dataset, the selected hyperparameters are $\beta = 0.7$, $T = 1.0$, learning rate $10^{-4}$, and 10 training epochs; smooth behavior across all four hyperparameters is shown in \cref{fig:scout_hyper}. These hyperparameters govern the $\hat{\mathbf{p}}$ model. The $\hat{z}$ model uses Ridge regression with regularization parameter $\alpha=10^{3}$, which similarly exhibits smooth behavior (\cref{fig:scout_alpha}).

\begin{figure}[h]
    \centering
    \begin{subfigure}[b]{0.49\linewidth}
        \centering
        \includegraphics[width=\linewidth]{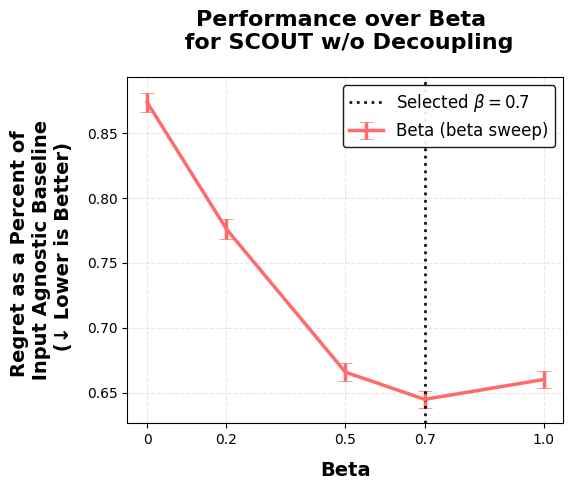}
        \caption{Beta sweep.}
    \end{subfigure}
    \hfill
    \begin{subfigure}[b]{0.49\linewidth}
        \centering
        \includegraphics[width=\linewidth]{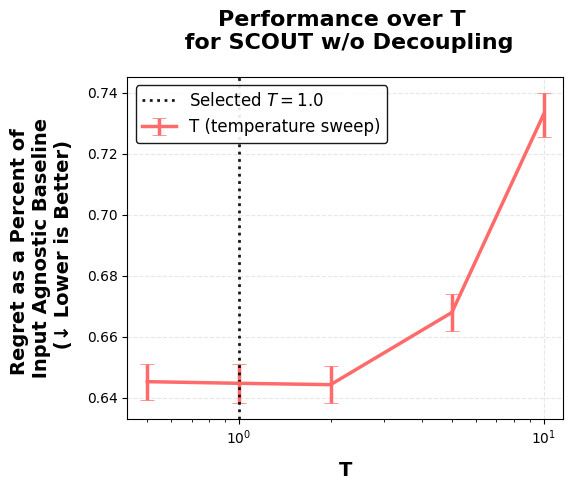}
        \caption{Temperature sweep.}
    \end{subfigure}

    \vspace{0.5em}

    \begin{subfigure}[b]{0.49\linewidth}
        \centering
        \includegraphics[width=\linewidth]{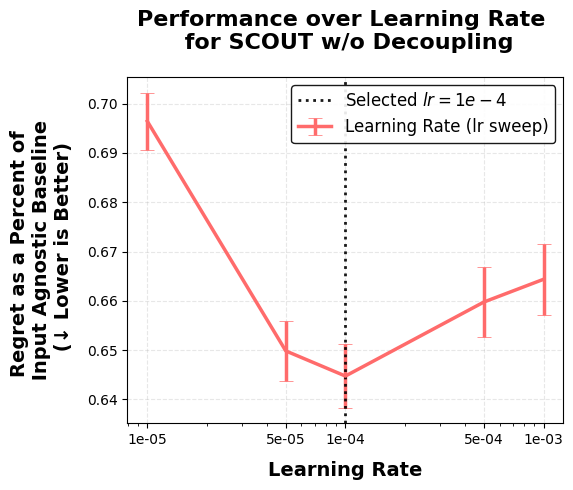}
        \caption{Learning rate sweep.}
    \end{subfigure}
    \hfill
    \begin{subfigure}[b]{0.49\linewidth}
        \centering
        \includegraphics[width=\linewidth]{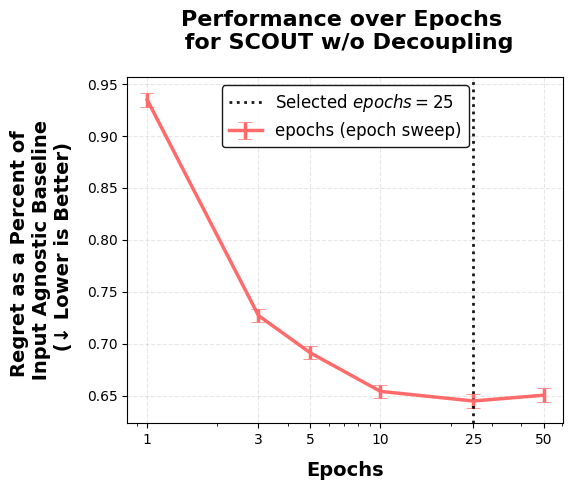}
        \caption{Epoch sweep.}
    \end{subfigure}
    \caption{Sensitivity of hyperparameters for SCOUT without decoupling evaluated on novel objects in the GSO dataset over the cost coefficient vector $\mathcal{C}_0$.}
    \label{fig:scoutnodc_hyper}
    \vspace{0px}
\end{figure}

We repeat this analysis for SCOUT without decoupling, using 25 epochs instead of 10, and find a similarly smooth landscape with the same remaining hyperparameters (\cref{fig:scoutnodc_hyper}). The increase in epochs is expected: without decoupling, the $\hat{\mathbf{p}}$ network must capture both relative model performance and overall image difficulty, requiring more training to converge.

\begin{figure}
    \centering
    \includegraphics[width=\linewidth]{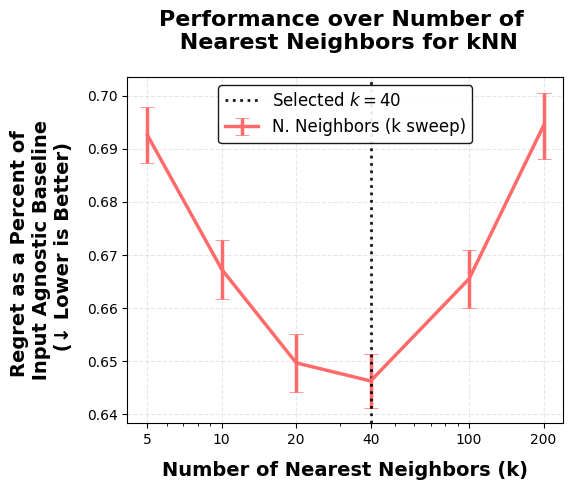}
    \caption{Sensitivity of the number of nearest neighbors for the kNN baseline evaluated on novel objects in the GSO dataset over the cost coefficient vector $\mathcal{C}_0$.}
    \label{fig:knnk}
    \vspace{0px}
\end{figure}

\begin{figure}
    \centering
    \includegraphics[width=\linewidth]{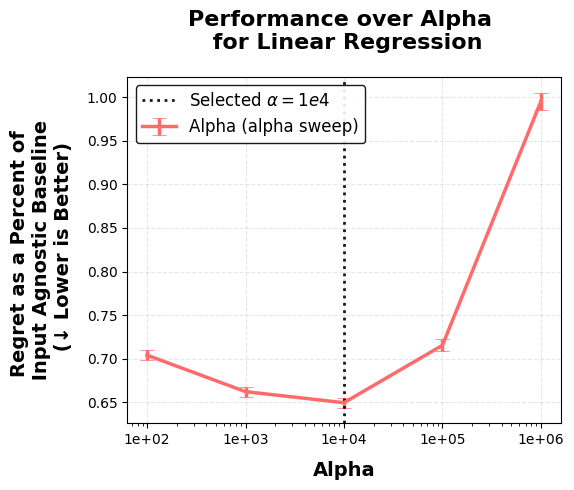}
    \caption{Sensitivity of alpha for the linear regression baseline evaluated on novel objects in the GSO dataset over the cost coefficient vector $\mathcal{C}_0$.}
    \label{fig:lralpha}
    \vspace{0px}
\end{figure}

\textbf{Baseline hyperparameters.} Since both our work and Li~\cite{li2025rethinking} find that kNN and LR models are among the strongest baselines, we analyze their hyperparameter sweeps as well. Both exhibit similarly smooth landscapes (kNN: \cref{fig:knnk}; LR: \cref{fig:lralpha}).

\section{Additional results}
\label{sec:additional_results}
\subsection{Weighted proxy results.}
\label{sec:weighted_proxy_results}
Our weighted proxy improves over all three alternatives---the ground-truth $z^{(i)}$, equal weighting, and one-hot weighting ($\tilde{z}^{(i)}=\tilde{z}_{j^*}^{(i)} \text{ where } j^*=\arg\max_jw_j$)---on the majority of metrics, as confirmed by one-sided t-tests (\cref{tab:ttests}). Most notably, it outperforms the ground-truth $z^{(i)}$ on 6 of 7 metrics with no deteriorations. The ground-truth $z^{(i)}$ is computed as a sum of exponentials over noisy reconstruction scores, so models with highly variable scores disproportionately inflate its variance. The derived weights mitigate this by downweighting model $j$ when (1) its predicted probabilities $\hat{p}_j$ are noisier (large $S_{j}^2(1-R_j^2)$), or (2) its true probability is frequently low, making the proxy $\tilde{z}_j^{(i)}$ a higher-variance estimate of $z^{(i)}$ (see \cref{eq:varzj}).

\begin{table}[h]
    \centering
    \caption{One-sided t-test $p$-values comparing our weighted partition function proxy against three alternatives on BigBIRD + YCB. Each cell reports the $p$-value for the hypothesis that our proxy achieves lower regret, averaged over cost coefficient vectors sampled from $\mathcal{C}$ (\cref{sec:experimentalsetup}). An improvement indicates a metric where our proxy achieves statistically significant ($\alpha = 0.05$) lower regret than the alternative; a deterioration ($\alpha = 0.05$ for the reverse one-sided test) indicates metrics where the alternative achieves statistically significant lower regret than ours.}
    
    \label{tab:ttests}
    \resizebox{\columnwidth}{!}{
    \begin{tabular}{|l|c|c|c|}
    
    \hline
    \textbf{Metric} & \textbf{True $z$} & \textbf{Equiweighted} & \textbf{One-hot} \\
    \hline
    DCD & ${3.24\text{e-}}4$ & $5.44\text{e-}1$ & ${2.38\text{e-}24}$ \\
    \hline
    Chamfer L2 & ${4.01\text{e-}35}$ & ${5.74\text{e-}35}$ & $8.11\text{e-}1$\\
    \hline
    Chamfer L1 & ${2.03\text{e-}4}$ & ${3.92\text{e-}2}$ & ${7.71\text{e-}16}$ \\
    \hline
    IoU & $4.35\text{e-}1$ & $2.31\text{e-}1$    & ${6.89\text{e-}5}$ \\
    \hline
    MMD-EMD & ${5.48\text{e-}13}$ & ${3.32\text{e-}6}$ & ${4.10\text{e-}4}$ \\
    \hline
    Eval3D-geo & ${8.24\text{e-}34}$ & ${2.70\text{e-}30}$ & ${9.09\text{e-}4}$ \\
    \hline
    Eval3D-struct & ${7.78\text{e-}18}$ & ${2.31\text{e-}18}$ & ${9.95\text{e-}1}$ \\
    \hline
    \hline
    $\#$ improvements & $6/7$ & $5/7$ & $5/7$ \\
    \hline
    $\#$ deteriorations & $0/7$ & $0/7$ & $1/7$ \\
    \hline
    \end{tabular}
    }
\end{table}
\subsection{Grasp collisions}
\label{sec:grasp_collisions}
\begin{figure*}
    \centering
    \includegraphics[width=\linewidth]{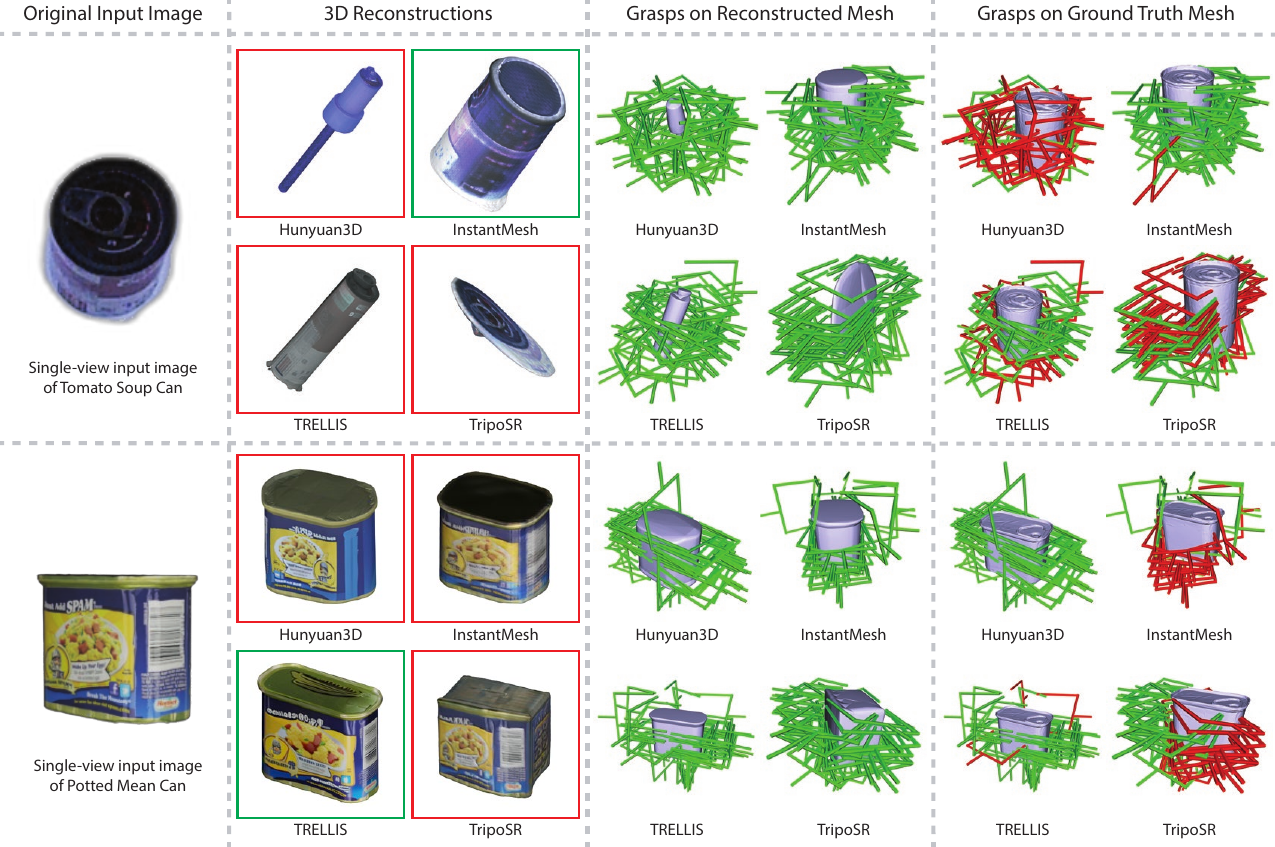}
    \caption{Evaluation of grasp proposals based on the reconstructed meshes from the input image. The figure shows the following from left to right: (1) original input image inputted to SCOUT, (2) 3D reconstructions from the original image with the model chosen by SCOUT boxed in green, (3) grasp proposals on the reconstructed mesh, (4) grasp proposals evaluated on the ground-truth mesh; colliding grasps are shown in red.}
    \label{fig:maingc}
    \vspace{0px}
\end{figure*}

\Cref{fig:maingc} illustrates two routing decisions chosen by SCOUT. For a top-down view of a can, only InstantMesh produces a usable reconstruction, and SCOUT selects it. For a frontal view of a meat container, SCOUT selects TRELLIS, which reconstructs the object accurately at a lower latency than Hunyuan3D. These examples show that per-input routing can improve reconstruction quality and reduce latency.

\begin{figure*}
    \centering
    \includegraphics[width=.9\linewidth]{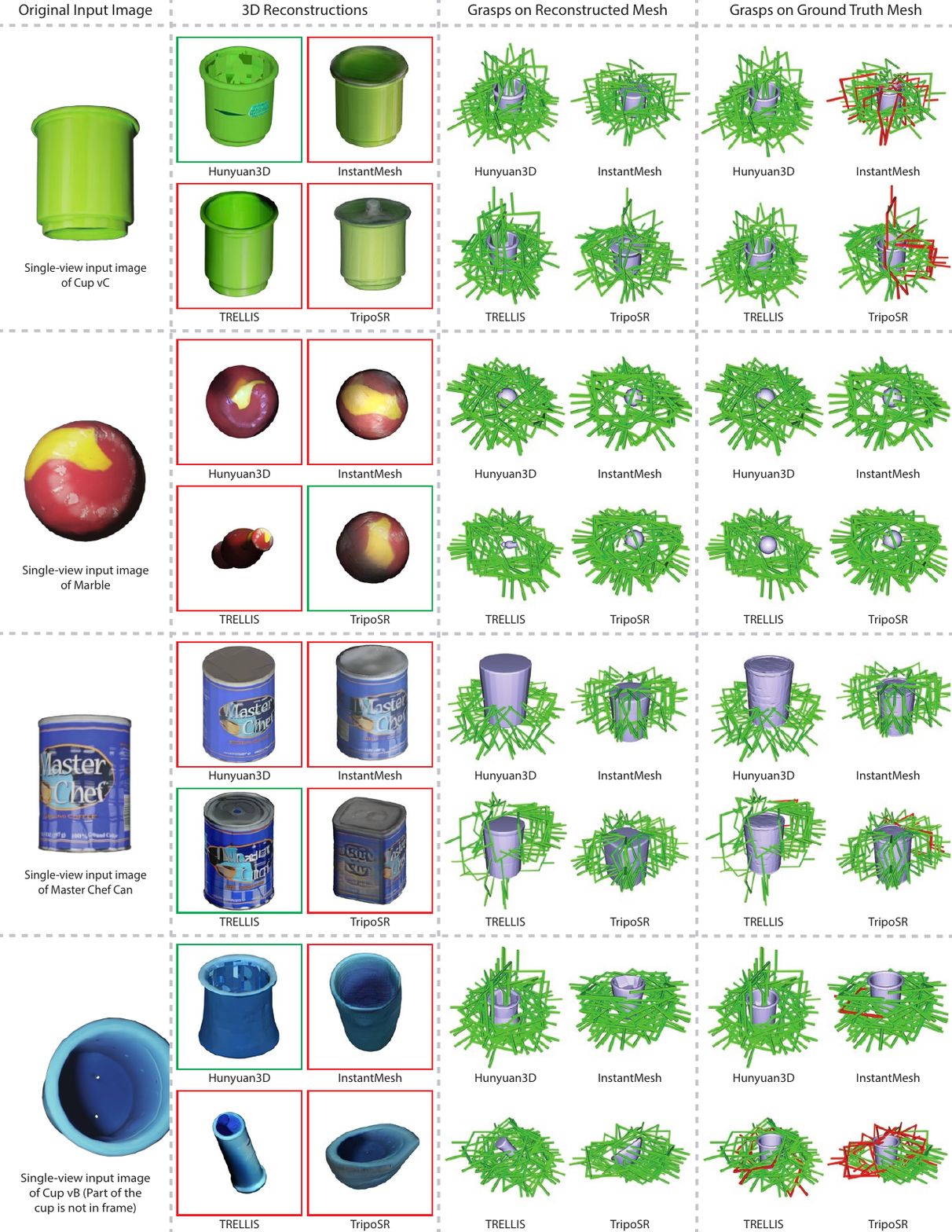}
    \caption{Evaluation of grasp proposals based on the reconstructed meshes from the input image. The figure shows the following from left to right: (1) original input image inputted to SCOUT, (2) 3D reconstructions from the original image with the model chosen by SCOUT boxed in green, (3) grasp proposals on the reconstructed mesh, (4) grasp proposals evaluated on the ground-truth mesh; colliding grasps are shown in red.}
    \label{fig:gcfirsthalf}
    \vspace{0px}
\end{figure*}

\begin{figure*}
    \centering
    \includegraphics[width=.9\linewidth]{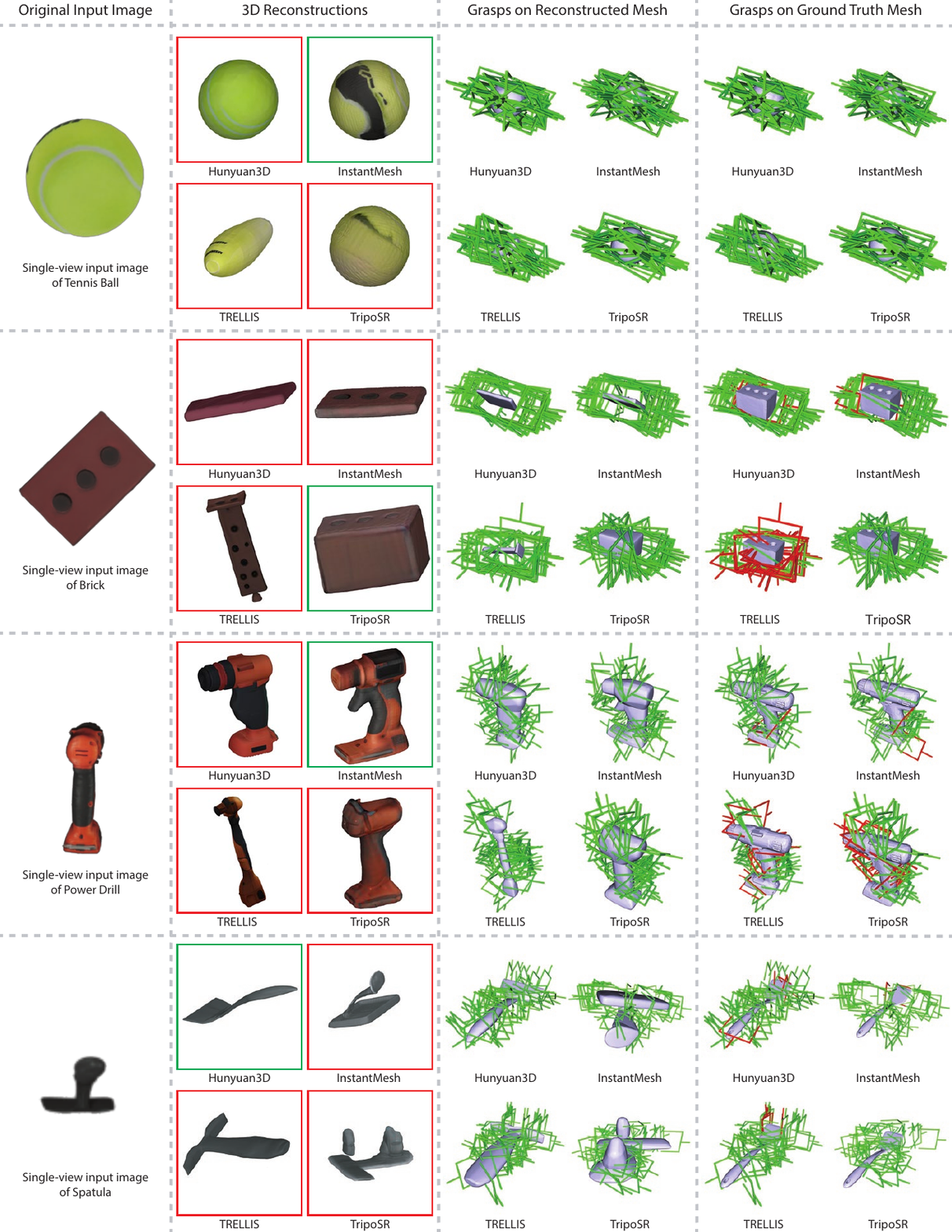}
    \caption{Evaluation of grasp proposals based on the reconstructed meshes from the input image. The figure shows the following from left to right: (1) original input image inputted to SCOUT, (2) 3D reconstructions from the original image with the model chosen by SCOUT boxed in green, (3) grasp proposals on the reconstructed mesh, (4) grasp proposals evaluated on the ground-truth mesh; colliding grasps are shown in red.}
    \label{fig:gcsecondhalf}
    \vspace{0px}
\end{figure*}

\Cref{fig:gcfirsthalf,fig:gcsecondhalf} show the eight remaining objects from the grasp collision experiment in \cref{tab:gc}. Across these objects, SCOUT routes to a variety of reconstruction models depending on the input viewpoint, confirming that no single model dominates and that per-input selection is beneficial.

\subsection{Policy learned}
\label{sec:policy_learned} 
\begin{figure}
    \centering
    \begin{subfigure}[b]{\linewidth}
        \centering
        \includegraphics[width=\linewidth]{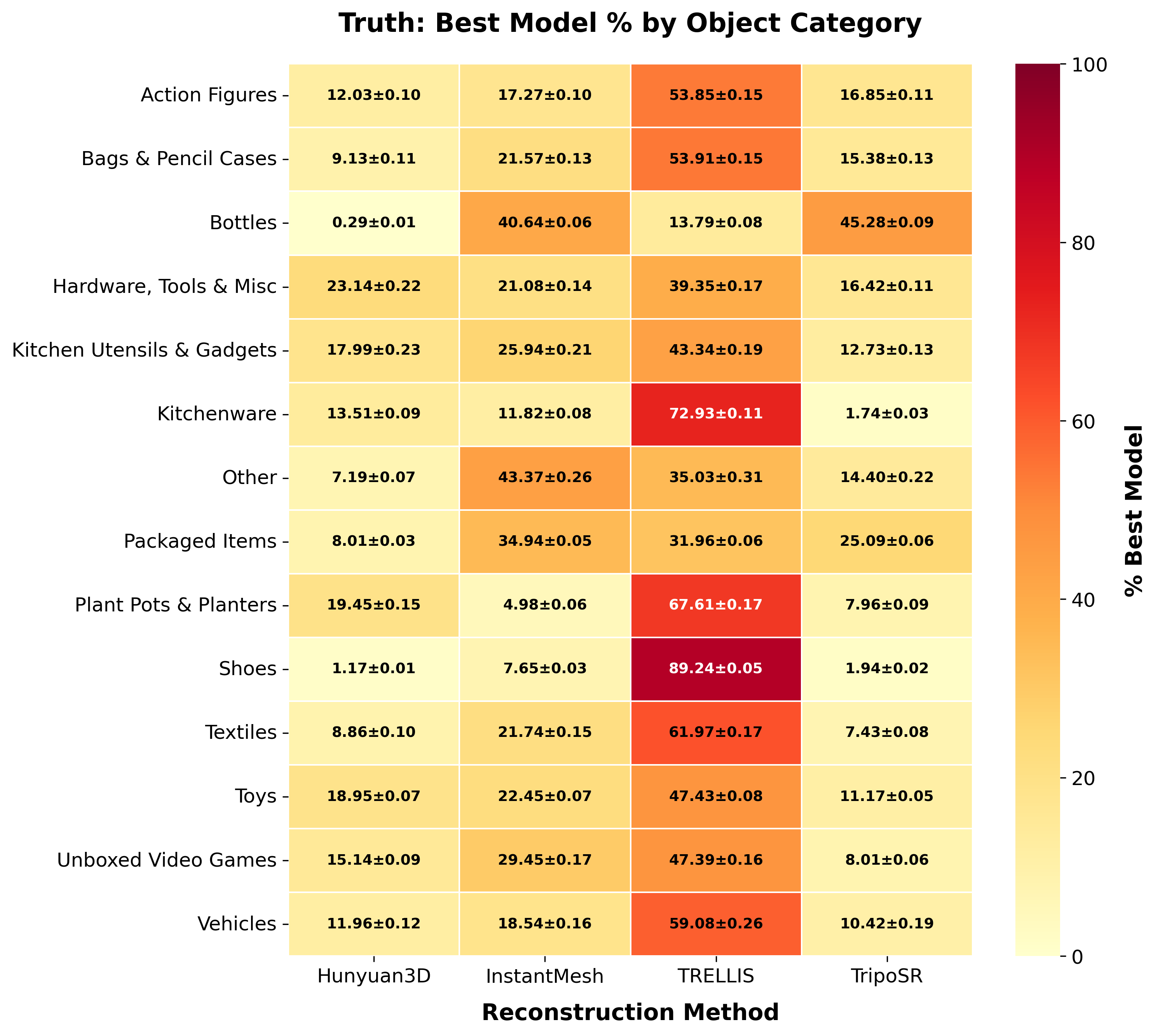}
        \caption{Oracle policy: how often each model is truly optimal when evaluated on unseen objects, grouped by object category.}
        \label{fig:categories_true}
    \end{subfigure}
    \vspace{0.5em}
    \begin{subfigure}[b]{\linewidth}
        \centering
        \includegraphics[width=\linewidth]{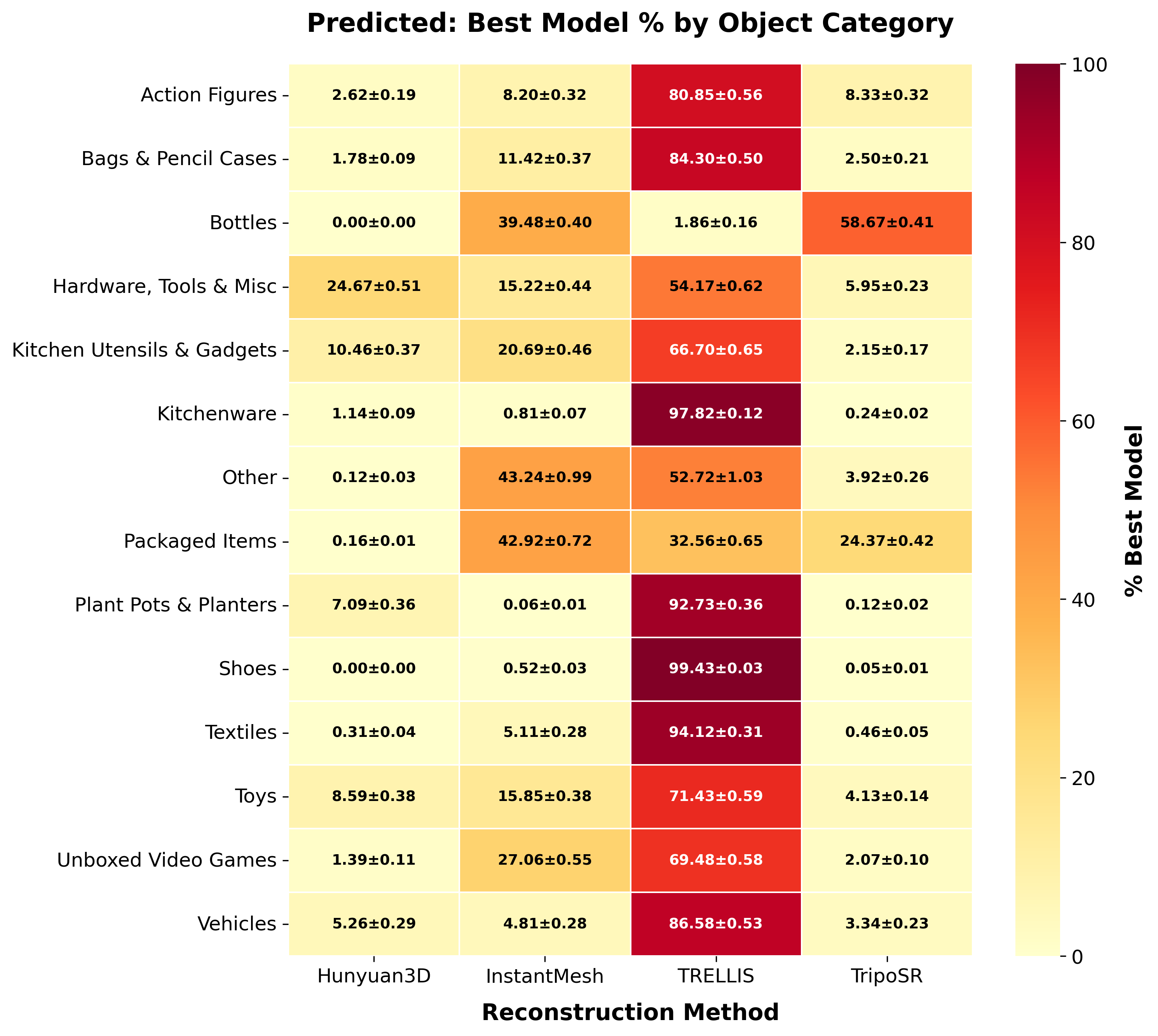}
        \caption{SCOUT's learned policy: how often each model is selected when evaluated on unseen objects, grouped by object category.}
        \label{fig:categories_pred}
    \end{subfigure}
    \caption{Oracle and learned routing policies across object categories.}
    \label{fig:categories}
    \vspace{0px}
\end{figure}

\begin{figure}
    \centering
    \begin{subfigure}[b]{\linewidth}
        \centering
        \includegraphics[width=\linewidth]{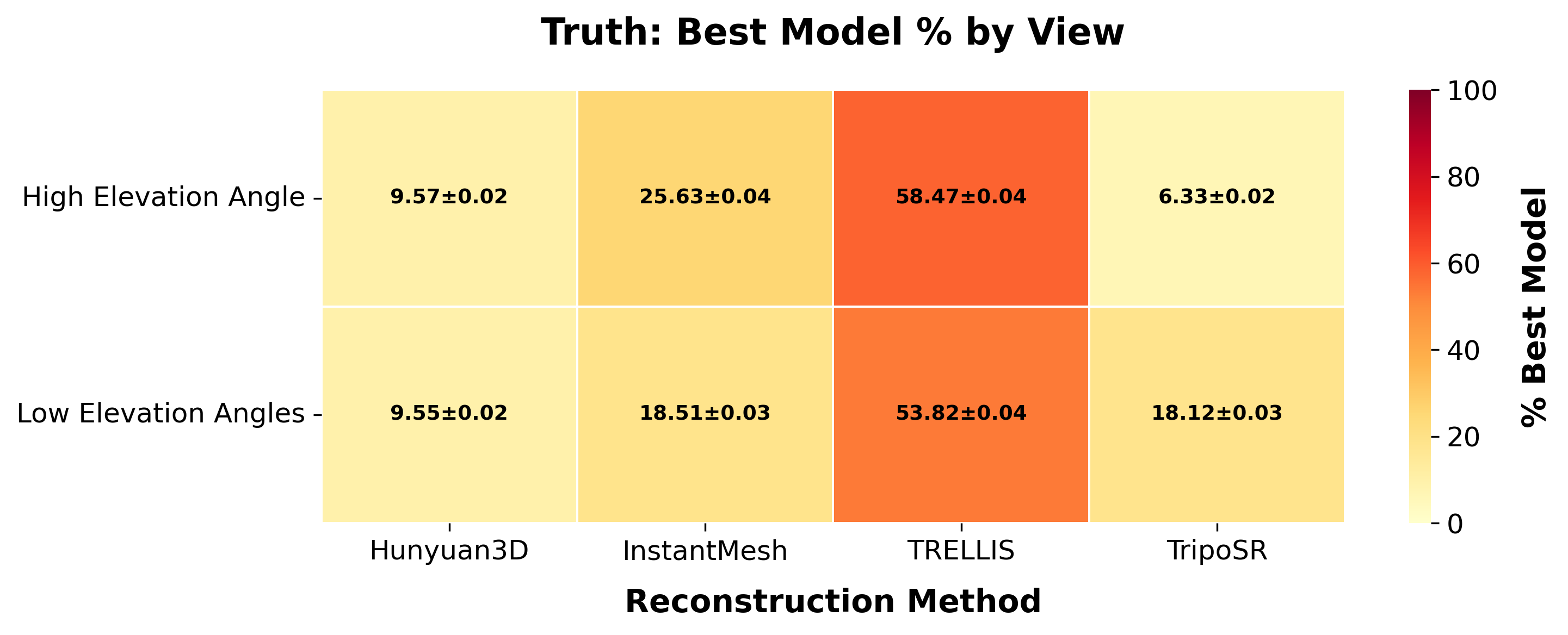}
        \caption{Oracle policy: how often each model is truly optimal when evaluated on unseen objects, grouped by elevation.}
        \label{fig:viewpoints_true}
    \end{subfigure}
    \vspace{0.5em}
    \begin{subfigure}[b]{\linewidth}
        \centering
        \includegraphics[width=\linewidth]{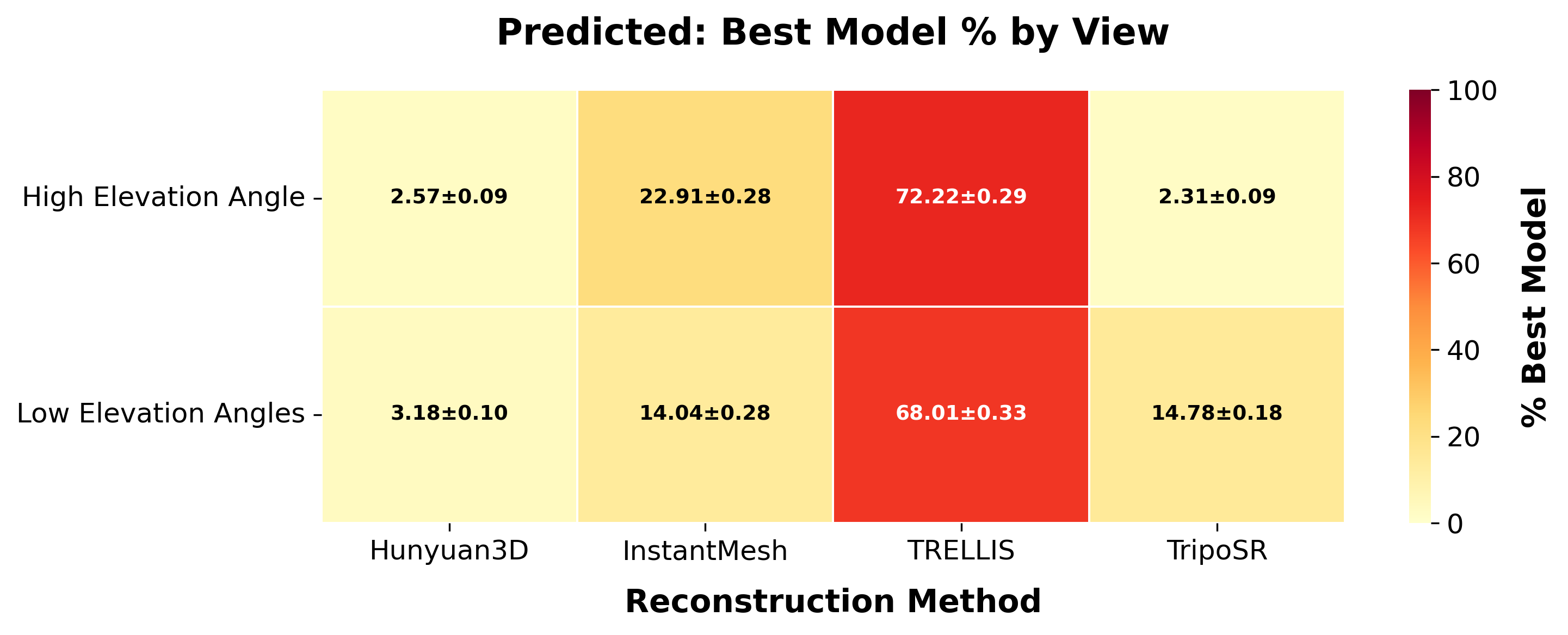}
        \caption{SCOUT's learned policy: how often each model is selected when evaluated on unseen objects, grouped by elevation.}
        \label{fig:viewpoints_pred}
    \end{subfigure}
    \caption{Oracle and learned routing policies across viewpoint elevations. High: $75^\circ$ and $85^\circ$; low: $35^\circ$, $45^\circ$, and $60^\circ$.}
    \label{fig:viewpoints}
    \vspace{0px}
\end{figure}

To understand what SCOUT learns, we examine which model is selected as a function of object category (\cref{fig:categories}) and input viewpoint (\cref{fig:viewpoints}). Different reconstruction models exhibit distinct biases across categories and viewpoints, and SCOUT's learned policy reflects these trends even when evaluating on novel objects, confirming that the router captures category-dependent biases and viewpoint-dependent biases from the training data.


\end{document}